%% file: main.tex
\documentclass[fleqn,10pt,twocolumn]{wlscirep}
\usepackage[utf8]{inputenc}
\usepackage[T1]{fontenc}
\usepackage{algorithm2e}
\usepackage{amsmath}
\usepackage{graphicx}
\usepackage{subcaption}
\usepackage{caption}
\usepackage[dvipsnames]{xcolor}
\definecolor{chunking}{RGB}{244,113,103}
\definecolor{rle}{RGB}{131,33,86}
\definecolor{pcfg}{RGB}{21,102,68}
\definecolor{ag}{RGB}{114,182,62}
\definecolor{hag}{RGB}{255,203,31}
\definecolor{chunking}{RGB}{0,0,0}
\definecolor{rle}{RGB}{0,0,0}
\definecolor{pcfg}{RGB}{0,0,0}
\definecolor{ag}{RGB}{0,0,0}
\definecolor{hag}{RGB}{0,0,0}

\newcommand{\lloc}{L_{\mathrm{loc}}}
\newcommand{\lglob}{L_{\mathrm{glob}}}
\newcommand{\aloc}{\alpha_{\mathrm{loc}}}
\newcommand{\aglob}{\alpha_{\mathrm{glob}}}
\newcommand{\dloc}{d_{\mathrm{loc}}}
\newcommand{\dglob}{d_{\mathrm{glob}}}

\title{Path-dependent program induction under resource constraints explains human sequence learning}

\author[1,2,3,4,*]{Hanqi Zhou}
\author[2,3,4]{David G. Nagy}
\author[1,4]{Peter Dayan}
\author[2,3,4]{Charley M. Wu}
\affil[1]{University of Tübingen, Tübingen, Germany}
\affil[2]{Center for Cognitive Science, Technical University Darmstadt, Darmstadt, Germany}
\affil[3]{Hessian.AI, Darmstadt, Germany}
\affil[4]{Department of Computational Neuroscience, Max Planck Institute for Biological Cybernetics, Tübingen, Germany}

\affil[*]{hanqi.zhou@uni-tuebingen.de}

\keywords{program induction, resource rationality, curriculum learning, sequence learning}

\begin{abstract} 
How do people build abstract, reusable knowledge from sequential experience under bounded cognitive resources?
To answer this question, we integrate rate-distortion theory with recent advances in program induction to describe how prior knowledge shapes which future structures are cheap to encode and easy to discover.
We formalize this in a hierarchical Adaptor Grammar (HAG) with distinct local (within-task) and global (across-task) libraries, governed jointly by constraints on memory and computation. 
In simulations, HAG achieves better rate-distortion trade-offs and stronger generalization than fixed grammars or shallow chunking methods. 
In an online melodic sequence-learning experiment, participants' recall errors reflected systematic simplifications and reaction times increased at inferred program boundaries. Trial-by-trial fits further showed that hierarchical libraries best explained individual differences in both recall and out-of-sample continuation choices, outperforming all alternative models. 
These findings cast structured learning as bounded program induction in which the order of experience shapes future abstractions a learner builds.

\end{abstract}
\begin{document}

\flushbottom
\maketitle
\thispagestyle{empty}

\section*{Introduction}
People do not merely memorize individual experiences. Instead, we distill noisy, sensory inputs into structured, reusable knowledge.
One influential account models this capacity as inference over the underlying generative processes that give rise to our observations, thus compressing noisy data into abstract concepts, rules, and skills~\cite{rule2020child, dehaene2022symbols, tenenbaum2011grow}. 
An emerging perspective formalizes this process as \emph{program induction}~\cite{lake2015human, goodman2014concepts}, whereby learners acquire compositional, language-like representations that support systematic generalization across a wide range of domains, including vision and shape perception~\cite{overlan2017learning,zhou2024compositional,tian2020learning}, language~\cite{piantadosi2008bayesian,piantadosi2016four,ellis2022synthesizing}, planning~\cite{sharma2021map,correa2024program}, music~\cite{wiggins2007compositional, harasim2020learnability,liu2016computational, nierhaus2009algorithmic} and abstract concept learning~\cite{zhao2023model, rule2024symbolic, dehaene2022symbols}. This can be seen as a concrete expression of the Language of Thought hypothesis~\cite{fodor1975language} (LoT).

However, program-based accounts of learning naturally raise the question: how are such representations shaped by the resource limitations under which human cognition operates? 
The principle of resource rationality formalizes cognition as optimization under constraints on time, memory, and computation, in order to identify the best achievable strategy (e.g., for encoding, decision-making) given those limits~\cite{lieder2020resource, simon1955behavioral}.

A key formal tool for analyzing a subset of these trade-offs is \emph{Rate-Distortion Theory} (RDT), which formalizes the balance between distortion (how much reconstruction error is incurred) and rate (how compactly a representation can be stored)~\cite{shannon1959coding, berger1971rate, gershman2021rational}. 
Although RDT has been applied to efficient perception~\cite{sims2016rate}, memory~\cite{nagy2020optimal, bates2020efficient}, and decision-making~\cite{bhui2018decision, lai2021policy}, two important limitations arise when using it as a framework for human learning. 
First, standard RDT assumes a fixed encoding scheme and characterizes optimal compression under that scheme. However, since a learner's representational repertoire changes over time, earlier experiences inevitably shape what can be efficiently encoded later, creating path-dependent dynamics~\cite{zhao2023model, dekker2022curriculum} that a static RDT cannot capture~\cite{nagy2025adaptive}. 
Second, RDT quantifies only the informational cost of storing a representation~\cite{gershman2021rational}, but is silent on the computational cost of discovering that representation in the first place. For structured hypothesis spaces such as programs, the search over candidate representations is combinatorially vast~\cite{franken2022algorithms}, and the effort required to find a good program is itself a major cognitive bottleneck. 

\input{figures/main/fig-main}

Thus, modeling human learning as resource-rational program induction requires extending classical RDT to address both gaps. First, we introduce a \emph{computational constraint} that limits the effort a learner can expend in proposing and revising candidate programs, formalizing the cost of inference. Second, we embed this trade-off within a learning process in which \emph{a growing library} of reusable programs reshapes both what is cheap to store and what is easy to discover, capturing the path-dependent dynamics.

Central to this framework is the idea of a \emph{generative library}, which holds a repertoire of reusable programs discovered and accumulated over time. 
Libraries help mitigate both constraints. Stored programs can be referenced by a compact pointer rather than spelled out in full, thereby reducing the description length (memory). Simultaneously, stored programs serve as high-quality proposals during inference, reducing the number of candidates that need to be considered and tweaked (computation). 
We further distinguish between a \emph{local library}, built within a task to capture context-specific regularities, and a \emph{global library}, which accumulates abstractions that generalize across tasks, together allowing hierarchical reuse of structure across multiple timescales.

Our account generates distinct predictions. First, learning should become progressively more compositional: as libraries grow, learners should increasingly construct new programs by combining previously acquired components, rather than encoding each sequence from scratch or memorizing verbatim (i.e., chunking). 
Second, learning should show specific path dependencies: once a learner has acquired a particular program, new inputs that share a program with the library should become cheaper to encode and easier to discover, whereas inputs that do not overlap with the current library should remain costly. Thus, the order in which regularities are encountered shapes which future abstractions are likely to be formed.
Together, these predictions imply characteristic behavioral signatures, including increased efficiency over learning, selective difficulty at inferred program boundaries, and systematic biases in recall and continuation toward simpler, reusable structure.

\subsection*{Goals and scope}
To test whether hierarchical program library induction explains how people acquire structured representations under cognitive constraints, we combine simulations with an online sequence-learning experiment using real-world melodies~\cite{garciavalencia2020sequence}.
Sequence learning offers a tractable setting for studying how learners infer latent structure from experience and reuse it to guide later encoding. The same surface sequence can be stored verbatim or represented more compactly as a generative program that captures higher-order regularities. These representational choices leave measurable behavioral signatures, which we test using behavioral and model-based analyses.

In simulations (Fig.~\ref{fig:main}a), we first demonstrate that hierarchical program induction offers more efficient compression and better generalization over existing alternatives, including non-adaptive grammars and shallow chunking accounts (Fig.~\ref{fig:main}c). 
We then assess it as a descriptive model of how human participants learn, recall, and predict continuations of sequences in an experiment (Fig.~\ref{fig:main}b).
Across measures, behavior aligned with the resource-rational predictions of our hierarchical library learning model: recall errors were shaped by systematic simplifications (consistent with compact programs under resource constraints) and reaction times (RTs) were selectively longer at model-predicted program boundaries.  
Finally, hierarchical library learning provided the best account of participants’ trial-by-trial choices, outperforming all alternatives, with the inferred libraries also making the best out-of-sample predictions of participants’ free-continuation choices. Together, these results support a view of learning as the joint product of program induction and library accumulation, driven by the construction of a compressed, generative library under bounded memory and computation.

\section*{Formalization}
We formalize sequence learning as inference over generative programs, supported by building a \emph{library} of reusable components. 
Learning is governed by two interacting constraints: a \emph{memory} constraint that penalizes representational cost of the inferred program (via description length) and a \emph{computational} constraint that bounds the ability to identify (search) and revise (backtracking) candidate programs.
To situate our account, we compare it with four alternative models that make progressively richer assumptions about structure (Fig.~\ref{fig:main}c; see Methods), from simple surface-level reuse to increasingly expressive structural representations.

First, \textcolor{rle}{tolerant Run-Length Encoding (RLE)} captures immediate repetitions as the simplest form of regularity by compressing consecutive identical symbols~\cite{golomb1966run}. Second,
\textcolor{chunking}{Chunking} extends note-level repetition to chunk-level by storing contiguous substrings as reusable units~\cite{miller1956magical, fonollosa2015learning, wu2023chunking}, even across sequences. However, both simply memorize and repeat previously observed strings, without forming richer abstractions.

In contrast, grammar-based methods capture richer structure by additionally supporting hierarchical composition.
Third, \textcolor{pcfg}{Probabilistic Context-Free Grammars (PCFGs)}~\cite{o2015productivity} are a formal approach for representing nested patterns through fixed production rules. While the probabilities of these rules can in principle be estimated from data, they are often specified in advance and fixed. Thus a standard PCFG does not provide a built-in mechanism for discovering and caching new reusable structures during inference (i.e., each sequence is parsed independently under the same grammar).
Fourth, \textcolor{ag}{Adaptor Grammars (AGs)} extend fixed grammars by adding experience-dependent reuse on top of a grammar. 
Intuitively, AGs allow for the caching and reuse of frequently used structures (i.e., programs) by maintaining a library~\cite{liang2010learning,zhao2023model}.
However, a key assumption is that the same library supports reuse across all contexts, neglecting the distinction between general purpose and locally relevant abstractions.

To extend AGs across different timescales of reuse, we introduce \textbf{Hierarchical Adaptor Grammars (HAGs)}, which maintains both a \emph{global} library to accumulate generic abstractions and a \emph{local} library for storing short-term patterns.
This hierarchical library lets HAG flexibly combine cross-task transfer with context-specific reuse, which fixed grammars (PCFG), surface-level repetition models (RLE/Chunking), and single-library AGs cannot capture.

Formally, to find an efficient representation for an observed sequence $X=\left[x_1, x_2, \ldots, x_T\right]$, HAG uses two interdependent processes, \emph{program inference} and \emph{library updating}. 
First, HAG searches for a generative program~$\Pi$ such that executing it produces a reconstruction~$\hat{X}=g(\Pi)$ of the observed sequence~$X$, where $g$ denote the interpreter that maps a program to its reconstruction. 
For later use, it is helpful to view a program~$\Pi$ as a concatenation (denoted as $\oplus$) of \emph{subprograms} $\Pi:=\pi_1 \oplus \pi_2 \oplus \cdots$, which are self-contained, executable subtrees that can be nested or concatenated to produce a subsequence. The reconstructed sequence is thus composed of several subsequences~$\hat{X}=[X_1, X_2, \ldots]$ where~$\hat{X}_i=g(\pi_i)$. 
After subprograms are selected, HAG updates two libraries: 
a local library~$\lloc$ that is built within the current sequence, and a global library~$\lglob$ that is updated across sequences to support transfer learning. 

\paragraph{Programs. }
We represent programs in Combinatory Logic (CL), as a minimal, variable-free and universal formalism for composing functions~\cite{liang2010learning}. 
Intuitively, CL provides a set of \emph{primitive operators} (``combinators'') that can be composed into binary trees, resulting in hierarchical expressions that generate sequences when executed. Any subtree is itself a complete subprogram that can be cached in the library, retrieved, and composed into new subprograms. 
In addition to standard CL combinators, we include predefined \emph{domain operators}, such that the same program language can express both compositional structure and common regularities in sequences (Fig.~\ref{fig:main}c; see Methods).
For example, $\texttt{up}(n, c)$~describes a program that ascends $c$~steps from note~$n$, such that the sequence~$D E F A$ can be concisely represented as~$\texttt{up}(D, 3)$. 

\paragraph{Program inference under dual constraints. }
Two constraints govern encoding and accumulation in program inference: \emph{memory} limitations on the amount of structure that can be stored, and \emph{computational} limitations on search over candidate programs and revision of earlier commitments. 

We first formalize \emph{memory} costs~\cite{gershman2021rational, nagy2025adaptive} as the representational complexity of a subprogram~$R_{\mathrm{mem}}(\pi)$. 
We use RDT~\cite{shannon1949mathematical, sims2016rate, zhou2024harmonizing} to characterize the goal of learning accurate representations (i.e., minimizing distortion) while also minimizing the complexity (i.e., rate). 
Thus, subprograms are selected to minimize the weighted sum of distortion and memory cost:
\begin{align}\label{eq:rdt-main-mem}
    \mathcal{L}(\pi)=d(X, \hat{X} \mid \pi) + \beta R_{\mathrm{mem}}\left(\pi \mid \lglob, \lloc \right) .
\end{align}
\noindent Here, $d(\cdot, \cdot)$~is the reconstruction error defined by Hamming distance and $\beta$~controls the trade-off between accuracy and complexity.
Importantly for our account, the memory cost~$R_{\mathrm{mem}}$ depends on the learner's current knowledge, since the description length of~$\pi$ depends on what the learner can already express compactly using~$\lglob$ and~$\lloc$. 
When a new sequence invokes a stored subprogram rather than re-deriving a new one, the description length is shorter. Thus, reuse helps reduce memory load through encoding recurring structure once and referencing it thereafter.

In addition to the cost of storing a representation~$R_\mathrm{mem}(\pi)$, we also constrain the \emph{computational} costs of discovering it $R_\mathrm{comp}(\pi)$. 
Minimizing the objective function in Eq.~\ref{eq:rdt-main-mem} requires searching a combinatorially vast space of possible subprograms. 
Thus we impose a soft limit on the inference process by replacing exact minimization with expected loss under a resource-bounded sampler $q_{\lambda_s, \lambda_b}$:
\begin{align}\label{eq:rdt-main-comp}
    \mathbb{E}_{\pi \sim q_{\lambda_s, \lambda_b}(\cdot \mid X)}[\mathcal{L}(\pi)] \,\, \text{subject to} \,\, R_{\mathrm{comp}}\left(\pi ; \lambda_s, \lambda_b\right) \leq C_{\text{limit}} .
\end{align}
\noindent The computational cost~$R_{\mathrm{comp}}$ can be decomposed into two budgets (Fig.~\ref{fig:main}a).
The \emph{search budget}~$N_s \sim \operatorname{Pois}\left(\lambda_s\right)+1$ limits how many candidate subprograms the learner can consider, while the \emph{backtracking budget}~$N_b \sim \operatorname{Pois}\left(\lambda_b\right)$ limits how far the learner can revise earlier commitments as new data arrives.
Crucially, because $q_{\lambda_s, \lambda_b}$ is stochastic, the learner is not guaranteed to find the loss-minimizing program. The realized representation reflects what is discoverable under the available budget (see Methods).

As an example, consider the learning trajectory in Figure~\ref{fig:main}a.
Upon observing the initial sequence~$D E F B$, the learner samples~$N_s$ candidate subprograms from its prior. It may provisionally accept a simple candidate such as~$\texttt{up}(D, 3)$, which predicts the sequence~$D E F G$ incurring distortion on the final note ($B \neq G$).
However, as new evidence arrives (e.g., the sequence continues~$B B B$), the learner infers that the first~$B$ may initiate a new regularity. Using its backtracking budget~$N_b$, it can revise the hypothesis to a sequentially composed subprogram, $\texttt{up}(D,3) \;\oplus\; \texttt{rep}(B,4)$.

\paragraph{Library sampling and updating. }
The local and global libraries define a structured prior over candidate subprograms using  a hierarchical Pitman-Yor process (HPYP; see Methods).
At each step, the learner samples a proposal that either reuses an element of $\lloc$, reuses an element of $\lglob$, or generates a novel subprogram from the base grammar:
\begin{align}
    p\left(\pi \mid \pi_{<j}^{(i)}\right) \propto \underbrace{n_{\text {loc }}(\pi)}_{\text {local reuse }}+\underbrace{\alpha_{\text {loc }} n_{\text {glob }}(\pi)}_{\text {global reuse }}+\underbrace{\alpha_{\text{loc}} \alpha_{\text{glob}} H(\pi)}_{\text {novel proposal }},
\end{align}
where $n_{\text{loc}}$ and $n_{\text{glob}}$ are usage counts in the local and global libraries, $H$ is the base grammar prior, and $\aloc, \aglob$ control the rates of falling back to the global library and the base grammar, respectively. 
After a subprogram is selected, its count $n_{\text{loc}}$ is incremented promoting further reuse within the current sequence. Once the sequence ends, a portion of these local counts propagates to $n_{\text {glob }}$, such that subprograms repeatedly useful within a sequence become more likely to be reused (as they are cheaper to represent and discover) in future ones. 
Together with the resource-bounded sampler, this yields path-dependent compression and transfer.

\section*{Results}
To show how dual constraints and the hypothesis of program induction jointly shape human sequence learning, we developed a sequence learning task using real-world MIDI melodies~\cite{garciavalencia2020sequence}. 
Both computational models and human participants observed melodic sequences and were tested on recall and generalization to new sequences, with performance quantified by reconstruction accuracy. 
We focus on melodic sequences because they exhibit cross-cultural universality~\cite{mehr2019universality}, and are characterized by rich~\cite{sievers2013music} and recursive structures~\cite{fitch2014hierarchical}. However, our framework applies broadly to any form of sequence learning.

\input{figures/simulation/rd-simulation/fig-rd-simulation}

\subsection*{Simulations}
We first report results from model simulations (Fig.~\ref{fig:rd-simulation}), which clarify how memory and computational limits jointly shape learning and generalization. 
When latent representations are structured as programs, our models achieve favorable rate–distortion (RD) trade-offs, especially with a hierarchical library (our HAG model). 
This benefit is consistent across both training and held-out generalization sets. Models with libraries also exhibit path-dependent learning dynamics---as training data accumulate, reusable programs are discovered and cached, leading to shorter description lengths and improved generalization to novel sequences. 

\paragraph{Compression performance. } 
To characterize the two trade-offs, we ran simulations in which each model inferred programs for sequences under different resource settings, allowing us to measure its rate-distortion (RD) behavior. Concretely, models were fit on a set of training sequences to update any learnable libraries, and then evaluated on held-out sequences to assess generalization (see Methods). 
Across $100$~simulation runs, we summarized performance by loss~$D+R$ at a fixed~$\beta=1$ (see SI for statistics and additional $\beta$ values). 

Across training sequences, HAG showed the most favorable RD trade-off: for a given rate, it had lower distortion than any alternatives, with AG typically second-best (Fig.~\ref{fig:rd-simulation}a, left). AG in turn outperformed RLE and Chunking, whereas PCFG performed worst overall. Thus, neither surface-level reuse nor fixed compositional structure alone matched the compression efficiency of learned hierarchical reuse.

This advantage also extends to held-out sequences. HAG continued to outperform all alternative models in generalizing to new melodies (Fig.~\ref{fig:rd-simulation}a, right), and we observed the same pattern in iterative optimization settings (Fig.~\ref{fig:si:eval-rd-curve}). These results suggest that HAG's learned global and local libraries support generalization by allowing unfamiliar sequences to be encoded with shorter description lengths at comparable distortion.
While these results are generally consistent across a broad range of parameters, a qualitatively different pattern emerges if we increase the memory constraint penalty to extreme values (i.e.,, $\beta=10.0$) that dominate the objective. With extreme memory constraints, all models prioritize reducing representational cost even when it increases distortion, such that the benefits of reusing structure diminishes and HAG’s performance drops to that of AG. 

\paragraph{Learning dynamics. } 
A core function of the library is to accumulate and reuse of programs, in order to make learning more efficient.
Consistent with this, all models with a library exhibited lower training loss over time ($D+R$ with~$\beta=1$; Fig.~\ref{fig:rd-simulation}b~left), whereas PCFG showed no improvement because it must relearn each sequence from scratch. 
Although all library-based models improve with training, they have different learning trajectories. Chunking improves rapidly early on with near-verbatim memory, resulting in low distortion but relatively high representational cost. But as the set of reusable substrings saturates, these improvements diminish. The same pattern holds in full RD curves under fixed computational budgets ($\lambda_s=10$, $\lambda_b=1$; Fig.~\ref{fig:si:rd-over-time}).

Beyond RD curves, a key prediction of library-based models is they can increasingly nest subprograms when constructing new ones. 
We quantify this via the average reconstruction length of subsequences generated by a single subprogram (Fig.~\ref{fig:rd-simulation}b, right). 
HAG subprograms consistently accounted for the longest subsequences under fixed computational budgets ($\lambda_s=10$, $\lambda_b=1$), followed by AG, while Chunking yielded substantially shorter subsequences. 
This pattern emerges most prominently when compression is weighted more strongly (i.e., $\beta$ is large). However, when $\beta$ is small (e.g., $\beta=0.1$; Fig.~\ref{fig:si:recon-len-per-program-trend-beta}), all models minimize distortion and therefore choose conservative subprograms that safely reconstruct shorter subsequences. In this regime HAG does not exceed AG or PCFG. But as $\beta$~increases, HAG and AG increasingly trade off small reconstruction errors for more economical descriptions by calling library items and composing them hierarchically, which yields longer subsequences.

\paragraph{Resource efficiency.} 
The previous analysis showed that HAG achieves the most favorable RD trade-off. We next asked how much computation is required to realize this advantage.
We therefore examined how performance varied with expected search budget $\lambda_s$ and expected backtracking budget $\lambda_b$ (Fig.~\ref{fig:rd-simulation}c; see Fig.~\ref{fig:si:rd-compute} for bivariate analyses).
Increasing the search~$\lambda_s$ yields only modest gains for HAG, AG, and RLE, especially later in training. Once a useful library has been learned, new sequences can be easily encoded by reusing a small set of general purpose subprograms, such that even shallow search often finds good candidates. 
Meanwhile, increasing the backtracking~$\lambda_b$ benefits all models by allowing earlier subsequences to be reinterpreted in light of later observations. For HAG and AG in particular, performance improves steadily with~$\lambda_b$, and AG closes much of the gap to HAG at high backtracking.

More broadly, search and backtracking play different roles. Increasing~$\lambda_s$ expands the pool of candidate subprograms and promotes the discovery of deeper compositional structure, whereas increasing~$\lambda_b$ enables revision of earlier representations, leading to longer reconstructed subsequences as partial solutions are progressively revised and extended. Figure~\ref{fig:si:computation-program} supports this interpretation by showing corresponding changes in program nesting depth and subsequence length.

\input{figures/behavior/error/fig-behavior-error}

Collectively, the simulations show that HAG outperforms others in compression efficiency and generalization, and that performance is bounded by memory and computation. These results motivate our human experiment of the same resource-bounded library-learning mechanisms: these constraints should shape both recall performance and learning dynamics, with library-based reuse governing experience-dependent generalization. We next test these predictions in an online experiment.

\subsection*{Experiment: three-phase melodic sequence learning} 

Guided by predictions from our simulations, we conducted an online experiment on Prolific with human participants ($n=96$; see Methods). 
Each participant was presented with five melodic sequences (sampled from the same dataset~\cite{garciavalencia2020sequence} as the simulations), with each sequence broken up into six segments (mean notes = 14.89) to ensure recall performance remains challenging but feasible (see Methods). 
The experiment consisted of three phases (Fig.~\ref{fig:main}b). 
In Phase~1, participants were asked to \emph{learn} each segment by reproducing it note by note with visual cues for the correct note. 
In Phase~2, participants were asked to \emph{recall} the segment from memory.
Finally, in Phase~3, participants were asked to \emph{predict} the next segment of notes based on what they previously observed.
Phases~1 and~2 were presented in alternating order for the first five segments, while the sixth segment was only used as the ground truth for Phase~3.

We report three sets of analyses that connect behavior to the program-based account. 
First, we examine Phase 2 recall errors and RTs for qualitative signatures of program-like compression, starting with model-agnostic descriptive analyses and then asking which mechanism  predicts choices and RTs well using a family of \emph{plausible observers}---models that generate theory-driven predictions across a broad grid of memory and computational parameter settings rather than being fit to behavior. This grid spans (and exceeds) the range later recovered from individual fits (Fig.~\ref{fig:si:fitted-param-distribution}), so the qualitative signatures we report do not depend on tuning, and the descriptive analyses are not biased toward more expressive models. 
Second, using the same observers, we test whether performance changes during learning reflects growth of internal libraries instead of generic practice. 
Finally, we fitted all models to trial-by-trial choices of individual participants and also evaluated out-of-sample generalization of Phase~3 predictions, testing whether our fitted libraries explain how people predict the continuation of a sequence beyond training.

\paragraph{Signatures of program-like compression. } 
We start by analyzing error patterns during the recall task in Phase~2 to reveal signatures of structured compression. 
Overall, participants achieved a mean recall error rate of~$.46 \pm .20$ (Fig.~\ref{fig:behavior-error}b; see Fig.~\ref{fig:si:error-rates-and-reaction-time}a for all three phases), which was lower than both a random baseline ($0.83$; Pop vs. Rand:~$z(95)=.0, p<.001,r=-1.0, \text{CI}=[-1.0, 1.0]$) and a first-order Markov model~$p(x_n \,|\, x_{n-1})$ fitted on training sequences ($0.64$; Pop vs. Stat:~$z(95)=.580, p<.001,r=-.75, \text{CI}=[-.87, -.61]$).

To first characterize the nature of recall error purely from behavior data, we classified note-level mismatches into four categories. 
Two categories served as diagnostic controls that can arise without structured representations: (i) \emph{vertical shift} are when a contiguous recalled subsequence differed from the target by a constant pitch offset, and (ii) \emph{temporal shift} are when a recalled subsequence matched the target after a lag. 
The other two categories were indicative of structure learning: (iii) \emph{repetition error} captured copying of earlier material, including both RLE-like repeated-note runs and chunk-like multi-note subsequences, and (iv) \emph{simpler program} captured cases in which recalled subsequence was better explained by a more compressible subprogram than verbatim memorization, but excluding repetition errors mentioned above (examples in Fig.~\ref{fig:behavior-error}a; see Methods). Together, these latter two categories test whether recall errors reflect systematic resource constraints, either through reuse or simplification.
Because errors can fall under multiple categories, we only counted unique errors assignable to a single category to avoid overcounting. 

Across participants, program-based simplifications were the most dominant category of errors ($.293 \pm .010$), followed by repeats ($.055 \pm .004$), temporal shifts ($.047 \pm .005$), and then vertical shifts ($.007 \pm .001$; Fig.~\ref{fig:behavior-error}c). 
Thus, participants not only exhibited lower overall error rates than chance, but also produced error patterns diagnostic of program-like compression.

We next asked which mechanisms best predict participants’ note-by-note recall choices, including both correct responses and systematic errors. 
To do so, we fit a logistic mixed-effects regression on participants’ choices using note-level expectation features for vertical and temporal shifts, together with predictors derived from RLE, Chunking, PCFG, AG, and HAG plausible observers. Intuitively, these predictors provide a score reflecting which choices were more likely under each model (see Methods). We jointly entered these predictors in a hierarchical softmax regression with participant-level random effects to predict each recalled note.
Overall, HAG had the largest positive coefficient (Fig.~\ref{fig:behavior-error}d; regression coefficient $\gamma = 5.82, z=68.62, p < .001, \text{CI}=[5.63, 5.98]$), indicating choices were systematically captured by assuming programmatic representations with reusable libraries. By contrast, AG, Chunking, and temporal shifts yielded much smaller positive effects, while RLE, PCFG, and vertical shifts were negative.
Excluding HAG significantly worsened fit, as demonstrated by a likelihood-ratio test comparing the full model to the nested model without the HAG predictor ($\chi^2(1)=2946.21, p<.001$). 

\input{figures/behavior/regression/fig-behavior-regression}

We then analyzed RTs for process-level evidence of program-like compression. 
In Phase~2, log-transformed RTs generally decreased across relative note position during a segment, but this decline was not monotonic: local jumps in RT duration were observed in addition to the overall downward trend, suggesting intermittent processing boundaries (Fig.~\ref{fig:behavior-regression}a, left).
To quantify these boundaries, we computed the note-to-note change in RT within each segment and marked a boundary when the change was unusually large relative to the participant and segment (thresholded at the mean $\pm$~1 s.d. of all changes within that segment). These boundaries divided segments into runs of relatively fast responses separated by longer pauses. The resulting ``chunk'' lengths averaged $4.81$~notes (Fig.~\ref{fig:behavior-regression}a, right), consistent with multi-note grouping rather than uniformly fast note-by-note responding (see Fig.~\ref{fig:si:error-rates-and-reaction-time}b for details).

We next asked whether these RT jumps align with subprogram boundaries. If participants internally represent sequences using concatenation of subprograms, initiating a new subprogram should incur a cognitive cost and produce an RT spike at the transition. 
To test this, we fit a mixed-effects regression that included model-predicted subprogram boundaries, along with note index as a control variable for generic decreases in RTs. 
Consistent with this prediction, HAG was the strongest predictor of increased RTs ($\gamma = 11.66, z=4.26, p < .001, \text{CI}=[6.30, 17.02]$) and significantly improved model fit when included as a regressor ($\chi^2(1)=18.18, p<.001$; Fig.~\ref{fig:behavior-regression}b). RLE was also a significantly positive predictor but smaller in magnitude ($\gamma = 5.66, z=11.18, p < .001, \text{CI}=[4.67, 6.65]$). 
However, AG and PCFG did not explain RT independently of HAG, likely because their boundary predictions overlapped with HAG (AG-HAG: $r=.66$; PCFG-HAG: $r=.58$); we therefore interpret their coefficients as reflecting collinearity rather than evidence against hierarchical structure. 
By contrast, Chunking had a negative coefficient, whereby predicted chunk boundaries were associated with faster responses. Because Chunking captures only short timescales of reuse, it predicts four times as many boundaries as HAG, most of which fall within what HAG treats as a single subsequence (see Fig.~\ref{fig:si:chunking-observer-rt}a-c and SI). This yields the seemingly paradoxical result that Chunking boundaries overall predicted faster responses.

\paragraph{Learning dynamics explained by libraries of subprograms. }
The preceding analyses showed that \emph{note-level} recall choices and RTs reflect signatures of program-like compression. We next asked whether \emph{segment-level} performance across the experiment can be explained by simple exposure-based practice alone, or whether it instead reflects changes in the internal libraries used to encode each segment. 

As a baseline, we first asked how much variance could be explained by practice alone (indexed by segment number and sequence number) without invoking any change in internal representations. Participants showed reliable practice effects: RTs sped up with both segments ($\gamma=-.031, z=-5.741, p<.001, \text{CI}=[-.041, -.020]$) and sequences ($\gamma=-.079, z=-15.377, p<.001, \text{CI}=[-.089, -.069]$). Consistent with these speed-ups, pauses became less frequent over learning, yielding longer contiguous runs (i.e., larger inferred chunks; Fig.~\ref{fig:si:learning-dynamics}a). Error rates also declined over learning ($\gamma=-.004, z=-3.555, p<.001, \text{CI}=[-.006,-.004]$ for segments; $\gamma=-.006, z=-5.742, p<.001, \text{CI}=[-.008, -.004]$ for sequences; Fig.~\ref{fig:si:learning-dynamics}b). However, these temporal indices explained relatively little variance in performance ($\mathrm{R}^2<.15$), indicating that simple practice effects capture only a limited portion of the learning dynamics.

We therefore asked whether segment-level performance was better explained by changes in the inferred representations. Applying all models as plausible observers to the same sequences participants experienced, we extracted for each segment a scalar estimate of \emph{segment program complexity}, as the description length of the best inferred program under the model's current library (see Methods). 
These complexity estimates were then used to predict segment-level recall error above and beyond the baseline practice indices. Segment complexity inferred by HAG ($\gamma=.211, z=8.068, p<.001, \text{CI}=[.160, .262]$), RLE ($\gamma=.138, z=6.550, p<.001, \text{CI}=[.089, .187]$) and PCFG ($\gamma=.343, z=5.623, p<.001, \text{CI}=[.084, .602]$) robustly predicted error rates, such that segments assigned higher program complexity were recalled less accurately (Fig.~\ref{fig:behavior-regression}c). 
By contrast, Chunking and AG were comparatively weak in the joint model because RLE, Chunking, and AG were highly correlated (Chunking vs. RLE $r=.65$, AG vs. RLE $r=.66$), indicating overlap in the complexity they capture if it comes down to a single scalar metric. Overall, adding program-based complexity predictors significantly improved fit relative to a model containing only temporal learning indices ($\chi^2(1)=24.11, p<.001$), indicating that learning is better explained by changes in inferred segment structure and library reuse than by time-on-task or surface complexity alone.

\paragraph{Model-fitting analysis}

\input{figures/model/fitting/fig-model-fitting}

Having demonstrated participant errors and RTs showed behavioral signatures of program induction under resource constraints, we next fitted each candidate model to individual participants' trial-by-trial choices during the Phase~2 recall task (see Methods). Model comparison was assessed using Bayesian Information Criterion (BIC) and hierarchical Bayesian model selection\cite{rigoux2014bayesian}. Additionally, we use fitted models to predict out-of-sample continuation data from Phase~3, which we evaluate using log loss.

Fitted to trial-by-trial choices of each participant, HAG had a significantly lower BIC (HAG vs. AG: $z(95)=227, p<.001, r=-.90, \text{CI}=[-.96, -.82]$).
To infer the most likely generative model at the population level, we conducted hierarchical Bayesian model selection by computing the protected exceedance probabilities (PXP). HAG yielded the highest PXP and the lowest BIC relative to all baseline models (Fig.~\ref{fig:model-fitting}a–b). 
At the individual level, HAG provided the best fit for $53$~of $96$ participants according to BIC, compared with $21$~for AG, $17$~for PCFG, $3$~for RLE, and $2$~for Chunking (Fig.~\ref{fig:si:model-fitting-score}).

We also evaluated model recovery to assess whether the fitting procedure could reliably identify the true generating model. All models were highly recoverable (Fig.~\ref{fig:si:model-recovery}), with the exception of AG ($P(\textrm{true}|\textrm{recovered})=.55$ in the inversion matrix), where AG-generated data was frequently best fit by PCFG, RLE, or Chunking.
AG implements a single, shared adaptor (a global library), which allows it to interpolate between qualitatively different behaviors depending on its adaptor strength ($\aglob$). In extreme regimes, weak reuse (higher~$\aglob$) yields behavior close to PCFGs, whereas in strong reuse (lower~$\aglob$), reuse can be amplified even when genuine sequence structure has not yet been identified. Early in training, this can lead the model to propagate whatever patterns are easiest to reuse---often shallow, high-frequency subsequences or repeats---which can resemble surface-level repetition models (RLE/Chunking) especially under finite data. 
By contrast, HAG has stronger identifiability ($P(\textrm{true}|\textrm{recovered})=.76$) and its parameters are recoverable (Kendall's rank correlation $\tau \ge .56$, all $p<.001$; Fig.~\ref{fig:si:param-recovery}). 
HAG's local library can capture within-sequence regularities without forcing them to generalize across sequences. As a result, sequence-specific patterns are less likely to be misrepresented as globally reusable structure, yielding clearer identifiability for both parameter and model recovery.

We then used our fitted models to make out-of-sample predictions of participant responses in Phase~3, where participants freely predicted the continuation of a sequence, extending what they observed in segments one to five.
For each participant, model fitting resulted in an individualized set of libraries inferred from their behavior up to the onset of Phase~3. 
We then treated the fitted libraries as a generative model to the sequence continuation, and compared these model generations to participants’ free responses. Because PCFG does not possess an explicit library representation, we treated it as a statistical baseline model, akin to a flexible $n$-gram model, and generated predictions using only its fitted parameters.
The results are shown in Figure~\ref{fig:model-fitting}c. 
At the aggregate level, HAG equipped with participant-specific libraries provided the best account of Phase~3 responses, with the lowest negative log-likelihood (nLL) among all models (HAG vs. AG: $z(95)=76381, p<.001, r=.323, \text{CI}=[.224, .420]$), and better than either a Markovian or chance baseline (${nLL}_\mathrm{HAG}=6.93$, ${nLL}_\mathrm{Stat.}=14.21$, and ${nLL}_\mathrm{Random}=25.69$) . 
Figure~\ref{fig:model-fitting}d shows two representative participants whose responses were closely predicted by HAG (blue represents match).

Beyond individual examples, we also assessed whether HAG captures systematic variation in Phase~3 responses at the population level. 
Across participants, the correlation between participant Phase~3 error rates and model-generated error rates are shown for each model (Fig.~\ref{fig:model-fitting}e). HAG had the strongest participant-level correspondence ($r=.23$, $p<.001$), exceeding AG ($r=.11$, $p=.070$), PCFG ($r=.06$, $p=.334$), and the simpler RLE ($r=.14$, $p<.05$) and Chunking baselines ($r=.12$, $p<.05$). Thus, beyond fitting individual continuations, HAG also best captured between-participant variability in how learners generalized to unseen continuations. Together, these results suggest that participant-specific hierarchical libraries support out-of-sample prediction of continuation behavior after limited exposure.

We next examined the fitted parameters of the winning HAG model.
For intuition, we report parameters such that larger values correspond to greater resource demands, facilitating a direct comparison of memory vs. computational trade-offs. Specifically, larger inverse temperature $\beta^{-1}$ reflects increased memory capacity, whereas larger values of the expected search budget ($\lambda_s$), expected backtracking budget ($\lambda_b$), and proposal rates ($\alpha, \aglob$) reflect increased computational capacity. 
Across participants, fitted parameters showed modest but systematic dependencies (Fig.~\ref{fig:si:model-recovery}c). Memory capacity ($\beta^{-1}$) was negatively correlated with computational capacity, including search budget ($\lambda_s$; $r=-.24$, $p<.01$), backtracking budget ($\lambda_b$; $r=-.20$, $p<.05$), indicating a trade-off between memory and computation. 
We also observed a strong positive correlation between the search budget ($\lambda_s$) and the tendency to propose new local programs ($\aloc$; $r=.19$, $p<.05$). This suggests that participants who searched more extensively also relied less on reuse of previously formed local structures and more on proposing novel candidates.

We next asked whether individual differences in parameter estimates could explain behavioral variability.
Participants’ error rates were significantly predicted by
both memory $\beta^{-1}$ ($\gamma=-.370$, $z=-141.32$, $p<.001$, $\text{CI}=[-.375, -.365]$) and computational capacity $\lambda_s + \lambda_b$ ($\gamma=-.051$, $z=-18.449$, $p<.001$, $\text{CI}=[-.027, -.022]$), with greater capacity yielding lower errors. 
In addition, computational resources positively predicted chunk size (Fig.~\ref{fig:model-fitting}c) derived from our RT analyses (Fig.~\ref{fig:behavior-regression}b), indicating greater computational capacity supports the formation of larger chunks ($\gamma=.038$, $z=24.478$, $p<.001$, $\text{CI}=[.035, .041]$).
To test whether these effects were captured quantitatively by the fitted models, we simulated each participant using their fitted parameters and compared predicted with observed individual differences in error rate and chunk size (Fig.~\ref{fig:si:simulation-learner}). For both measures, HAG enjoyed the strongest participant-level correspondence and explained unique variance in joint mixed-effects regressions when controlling for competing models (see Fig.~\ref{fig:si:simulation-learner} for details).

Finally, we examined parameters governing global versus local reuse. We used the parameter contrast $\aglob-\aloc$ as an index of relative reliance on across-sequence versus within-sequence reuse, and quantified each participant’s training materials using a similarity contrast comparing structure shared across sequences to structure repeated within sequences. Concretely, this similarity contrast summarizes the extent to which a participant’s five training sequences contain recurring motifs and transformations that reappear across different sequences versus what is confined to individual sequences.
The similarity contrast significantly predicted the reuse-parameter contrast: participants exposed to training sets with more cross-sequence relative to within-sequence structure showed larger values of $\aglob-\aloc$ ($\gamma=.277$, $z=102.254$, $p<.001$, $\text{CI}=[.271,.282]$), indicating greater inferred reliance on a global library. This links individual reuse strategies to the structural statistics of the training sequences participants experienced.

\section*{Discussion}
We asked how humans acquire structured representations of sequential data when memory and computation are limited, and how these representations change with experience. 
Integrating the Language of Thought (LoT) hypothesis~\cite{fodor1975language} with principles of resource rationality~\cite{simon1955behavioral,lieder2020resource}, we developed a new formalization of learning as path-dependent program induction, under joint constraints of memory and computation. 
We tested this framework against existing accounts---ranging from surface-level chunking to fixed grammars---using both simulations and a behavioral experiment in which participants learned, recalled, and predicted continuations of melodic sequences. 
The results all converge on a view in which learners compress sequences into compositional, program-like representations while trading accuracy off against two cognitive costs. The first is a \emph{memory} cost, determined by the description length of a sequence relative to an evolving library of reusable structure. The second is a \emph{computational} cost, determined by the bounded search and revision required to discover a candidate representation in the first place.

Existing accounts of structure learning~\cite{kemp2008discovery} typically address these elements in isolation. Program induction and library learning specify expressive hypothesis spaces but leave resource costs implicit~\cite{lake2015human,rule2018learning,ellis2021dreamcoder}. Rate-distortion approaches formalize the trade-off between fidelity and representational cost, but assume fixed representations which are optimized for a predefined training distribution~\cite{gershman2021rational}. 
Our framework unifies these components, recasting structure learning as resource-constrained program induction with an explicit theory of computation.
In doing so, it shifts the central question from what patterns are representable in principle, to which abstractions a learner will actually converge on, given their memory limits, computational budget, and the path-dependent history of structures they have previously experienced.

Our empirical results anchor this claim across fours converging lines of evidence---recall errors, reaction times, performance improvements, and sequence continuations---each of which connects to distinct predictions of the program-compression account.
First, participants' recall errors were not captured by assuming they memorized isolated notes or by shallow transition statistics. Instead, errors were systematically structured, often simplifying sequences into more compressible forms (Fig.~\ref{fig:behavior-error}c). 
This pattern follows from the compression objective (Eq.~\ref{eq:rdt-main-comp}): under resource constraints, learners favor representations with lower description length, even when they distort the original sequence. Errors therefore cluster toward ``nearby'' programs in the hypothesis space---sequences that share structural primitives with the target but require fewer bits to encode~\cite{brady2009compression,bates2020efficient}. This predicts the predominance of regularization errors (e.g., converting irregular patterns to exact repetitions)~\cite{reali2009evolution,culbertson2012learning} over arbitrary substitutions, consistent with prior work on memory distortions toward schematic prototypes and rule-based simplifications~\cite{hemmer2009bayesian,reali2009evolution, nagy2025adaptive}.
 
Second, RTs showed process-level signatures of latent program structure, with participants slowing at boundaries predicted by inferred subprograms. 
When learners encode sequences as programs, transitioning from one subprogram to the next should incur a cognitive cost---analogous to the initiation costs observed in hierarchical action control~\cite{lashley1951problem,rosenbaum2007problem,sakai2003chunking}, and music performance~\cite{verwey1996practicing}. HAG boundary predictions were the strongest predictor of these RT increases, consistent with the view that participants internally segmented sequences at the joints of compositional subprograms rather than at surface-level repetition boundaries.
 
Third, performance improvements over the course of learning were better explained by changes in inferred program complexity than by time-on-task alone. Segments assigned higher description length under the model's evolving library were recalled less accurately, even after controlling for surface statistics such as sequence length and transition entropy. Improvements with experience thus reflect accumulated structural knowledge (i.e., library learning): once a useful program has been acquired, later sequences that overlap with it become both cheaper to encode and easier to recover under bounded search~\cite{ellis2021dreamcoder,xia2021temporal}.
 
Fourth, these inferred libraries also generalized beyond the training material. In the prediction phase, participants recombined previously acquired structure to predict the novel continuation of a sequence.  Participant-specific HAG libraries predicted this behavior better than all competing models. 
Thus, the same libraries that best fit each participant's recall behavior also best predicted the open-ended sequences they later produced, indicating that the same representations support both encoding of observed structure and recombination of acquired structure to generate novel outputs. 
 
At the individual level, the estimated HAG parameters revealed structure as to how participants allocated their cognitive resources, going beyond the expected relationship between capacity and accuracy.
First, RT-inferred chunk sizes were larger for participants with larger estimated search and backtracking budgets. A purely capacity-based account---a bigger buffer licensing bigger chunks---would also predict this correlation. 
However, the chunk sizes we recovered from our RT analysis reflect the \emph{output} length of the inferred programs from our HAG model rather than the length of memorized subsequences based on a Chunking model. Additionally, chunk sizes were correlated with the estimated computational budget, suggesting computation was being spent on inferring such programs rather than simply stitching stored experiences together.
Second, the estimated reuse parameters were shaped by the statistical structure of what participants experienced. The contrast $\aglob - \aloc$, indexing relative reliance on cross- versus within-sequence reuse, increased with a stimulus-derived similarity contrast quantifying how much structure recurred across a participant's training sequences versus within individual sequences. Each learner's allocation between the two libraries tracked the regularities they had actually encountered, suggesting a form of meta-control that adjusts where to invest reuse based on the input. The two libraries compete on different cost profiles: the local library is cheap to populate within a sequence but resets at sequence boundaries, whereas the global library is slower to accumulate abstractions across sequences but persists and supports transfer. This meta-control is itself bounded by capacity, since under a tight memory budget, the global library cannot be large. 
Together, these findings show that individual differences are not reducible to capacity alone. They reflect the joint contribution of three factors: how much memory and computation a learner has, how that capacity is split between local and global reuse, and how prior experience shapes which structures are cheap to reuse going forward.

This sensitivity to prior experience generalizes beyond individual variation, providing a principled basis for path dependence in learning more broadly~\cite{ritter2007order,bruner1973organization,dayan2020first}. Because libraries reshape both what is cheap to encode and what is easy to infer, early experiences determine which structures enter the library, and those early commitments bias which regularities are easily discoverable under bounded search. 
Representations are more salient when they share structures with prior experience, even when surface details are new.
As learners accumulate reusable programs across sequences, acquired libraries progressively change both the representational and computational landscape of future learning. Path dependence thus follows from library-based compression as a structural prediction, with quantitative specificity---the model commits to which later sequences should be cheap to encode (those sharing subprograms with the current library) and which should not. 
What a learner ends up knowing depends as much on the order of their experience as on its content---a long-standing intuition our framework now makes precise: given a participant's history, the libraries we infer turn path dependence from an observation into a prediction.

Of course, our conclusions are subject to some limitations.
Music has long served as a testbed for theories of structured cognition, supported by rich traditions of work, e.g., meter and rhythm \cite{lerdahl2001tonal, london2012hearing}, expressive timing \cite{palmer1997music, repp1995quantitative}, long-range melodic and harmonic dependencies \cite{pearce2018statistical, harasim2020learnability}. 
Our aim was not to contribute a music-specific account to this literature, but to use melodies as an ecologically valid setting in which compositional and recursive structure is rich yet tractable. The framework we develop is therefore deliberately generic, formulated over arbitrary symbolic sequences. A fuller account of music learning would naturally need to integrate it with the dimensions that domain-specific models capture. 
By the same token, this generality is also a strength. The same resource-rational program-induction approach should apply to other naturalistic domains in which compositional reuse and abstraction are central, including language and mathematics~\cite{ellis2021dreamcoder, zhou2024predictive}, offering a unified test of whether library learning is a shared mechanism for structured knowledge acquisition.

A second limitation concerns the scope of resource costs we model. Although our framework adds search and backtracking alongside standard informational costs, it still does not exhaust the space of computational constraints. 
Other potential costs, such as maintaining, updating, or refactoring the learned libraries~\cite{ellis2021dreamcoder, bowers2023top}, remain outside the current formulation. By bringing inferential cost into the same framework as memory cost, our account broadens the resource-rational toolkit for studying structured learning, and points toward further extensions in which library curation itself is treated as a costly process.

Future work can test the predictions of path-dependent program induction with targeted manipulations. Since the framework predicts that learning is shaped by the order in which structure is encountered, systematically varying the overlap and sequencing of training materials should produce curriculum effects: learners who encounter foundational subprograms early should build richer libraries and encode later sequences more efficiently than those who encounter the same material in a less scaffolded order \cite{dekker2022curriculum, zhao2023model}. 
Orthogonal to learning order, manipulations of computational budget (e.g., time pressure or dual-task load) would provide a complementary test. Recent work shows that time pressure does not simply degrade accuracy uniformly but instead amplifies a bias toward compositional reuse of previously acquired structure \cite{rubino2026simplicity}, which is consistent with our framework's prediction that tighter computational budgets push learners to lean more heavily on existing library entries rather than search for new ones.

In sum, the findings support a view of sequence learning in which people construct hierarchical, reusable programs under joint memory and computational constraints. On this account, distortions, boundary costs, transfer, and individual differences all reflect the same adaptive process, whereby structured representations are built through lossy compression, and learning changes the space of future hypotheses by curating a library of abstractions available for reuse.

\input{method.tex}

\section*{Acknowledgments}
We thank Bonan Zhao for sharing code that informed our implementation, Josh Rule for helpful discussion, Aswath Chandrasekaran for implementing the experiment interface, and audiences at CogSci 2025 for valuable feedback. 
HZ thanks the International Max Planck Research School for Intelligent Systems (IMPRS-IS) for support. PD was supported by the Max Planck Society and the Humboldt Foundation.
This work is supported by the European Research Council (ERC) under the European Union’s Horizon 2020 research and innovation programme ($C^4$: 101164709), the Hessian research funding programme LOEWE/4b//519/05/01.002(0022)/119, the Deutsche Forschungsgemeinschaft (German Research Foundation, DFG) under Germany’s Excellence Strategy (EXC 3066/1 ``The Adaptive Mind'', Project No. 533717223), and the Excellence Cluster ``Reasonable AI'' by the Deutsche Forschungsgemeinschaft (German Research Foundation, DFG) under Germany’s Excellence Strategy – EXC-3057. 

\section*{Author contributions statement}

H.Z., D.G.N., P.D., and C.M.W. conceived the simulations and experiments, H.Z. conducted the simulations, H.Z. and C.M.W. conducted the online experiments, H.Z. analysed the results. H.Z. wrote the first draft of the
manuscript with feedback from D.G.N., P.D., and C.M.W. 
All authors reviewed and approved the final manuscript.

\section*{Additional information}

Data and code are publicly available at \href{https://github.com/zhouhq10/program-induction-rdt}{github.com/zhouhq10/program-induction-rdt}. 
The authors declare no competing interests.

\bibliography{reference}

\clearpage
\onecolumn
\renewcommand{\thefigure}{S\arabic{figure}}
\setcounter{figure}{0}
\renewcommand{\thetable}{S\arabic{table}} 
\setcounter{table}{0}
\renewcommand{\thealgocf}{S\arabic{algocf}}
\setcounter{algocf}{0}

\include{si}

\end{document}

%% file: figures/main/fig-main.tex
\begin{figure*}[h] 
    \centering
    \includegraphics[width=0.95\linewidth]{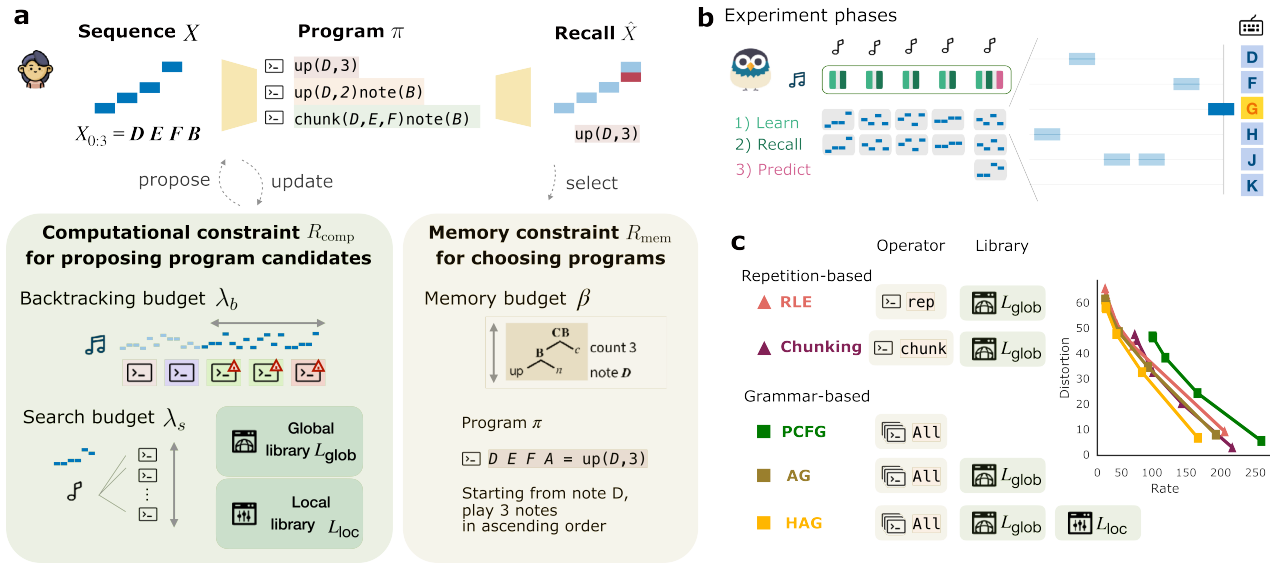}
    \caption{
    \textbf{Overview. }
    \textbf{(a)} \emph{Hierarchical Adaptor Grammar} (HAG) model describes sequence learning as inference over programs under dual resource constraints. 
    An observed sequence~$X$ (left) is encoded as a generative program $\pi$. A \emph{memory} constraint sets the representational budget (i.e., rate) for the final program, with~$\beta$ controlling how much shorter programs are preferred over minimizing distortion. A \emph{computational} constraint limits inference by bounding the number of candidate programs that can be proposed and revised, via a search budget~$\lambda_s$ and a corrective backtracking budget~$\lambda_b$. Candidate programs draw on a global library~$L_{\mathrm{glob}}$ and a local library~$L_{\mathrm{loc}}$, which are updated as structure is discovered.
    \textbf{(b)} Melody-learning experiment. Participants (and model simulations) (1) learn a melodic sequence note-by-note, (2) recall from memory, and (3) predict out-of-sample continuations to probe generalization.
    \textbf{(c)} Simulated rate-distortion curves. We contrast repetition-based (\textcolor{rle}{RLE}, \textcolor{chunking}{Chunking}) and grammar-based models(\textcolor{pcfg}{PCFG}, \textcolor{ag}{AG}, \textcolor{hag}{HAG}). Icons indicate which primitive operators (repetition/chunking/all) and library components (global/local) is used in each model. Curves show the trade-off between representational rate and distortion.
    }
    \label{fig:main}
    \vspace{-1em}
\end{figure*}

%% file: figures/simulation/rd-simulation/fig-rd-simulation.tex
\begin{figure}[!t]
    \centering
    \includegraphics[width=3.2in]{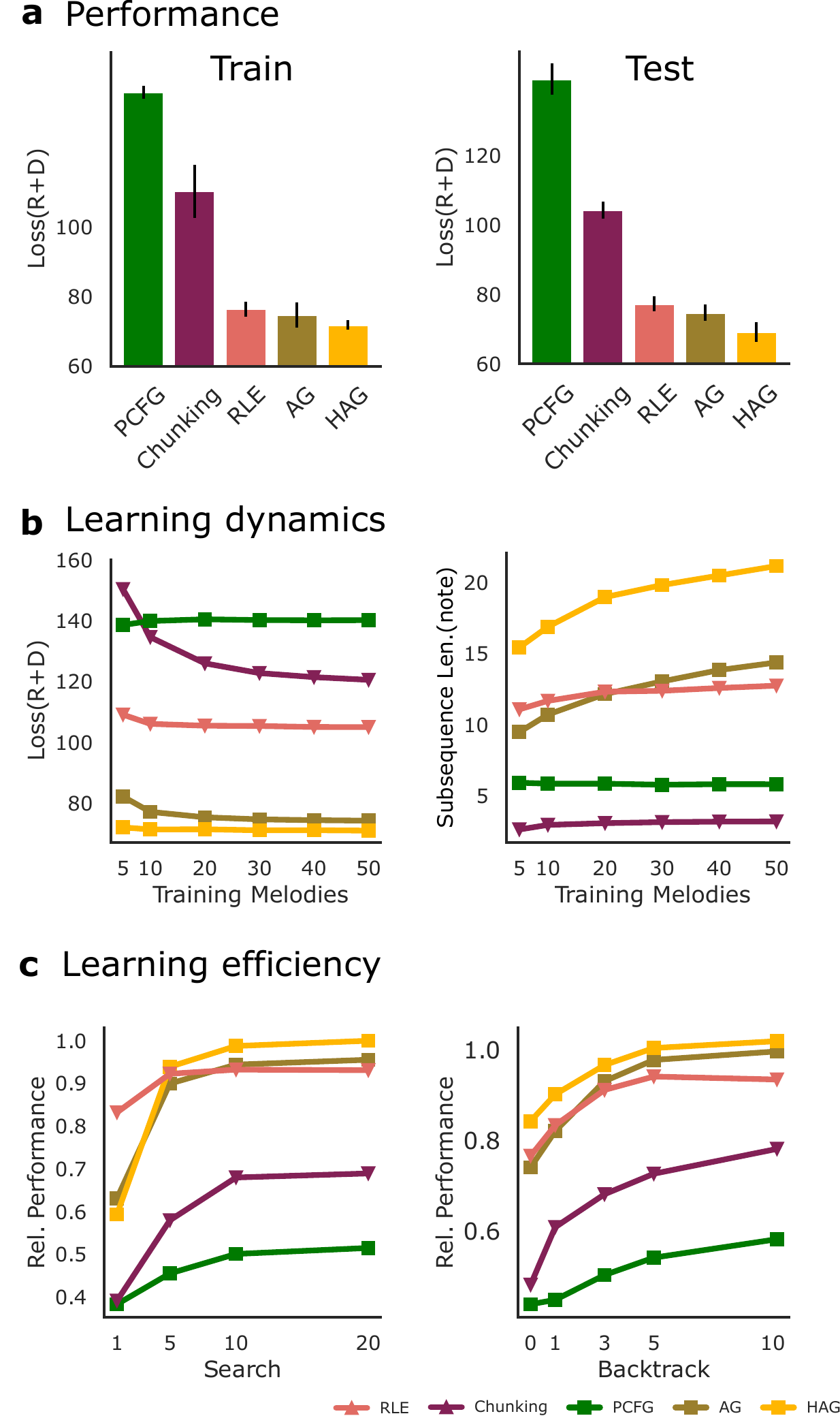}
    \caption{\textbf{Simulation results for the melodic sequence learning task.}
    We study how memory (rate), computation (search and backtracking), and accuracy (distortion) interact over the course of learning. 
    \textbf{(a)} Loss objectives (rate + distortion) for training sequences and one-shot generalization to held-out sequences using learned libraries.
    \textbf{(b)} Evolution of loss and subsequence length per subprogram as training data increases. 
    \textbf{(c)} RD values as a function of computational resources, showing how search budget ($\lambda_s$) and backtracking budget ($\lambda_b$) affect model performance.}
    \label{fig:rd-simulation}
    \vspace{-1em}
\end{figure}

%% file: figures/behavior/error/fig-behavior-error.tex
\begin{figure}[!ht]
    \centering
    \includegraphics[width=0.95\linewidth]{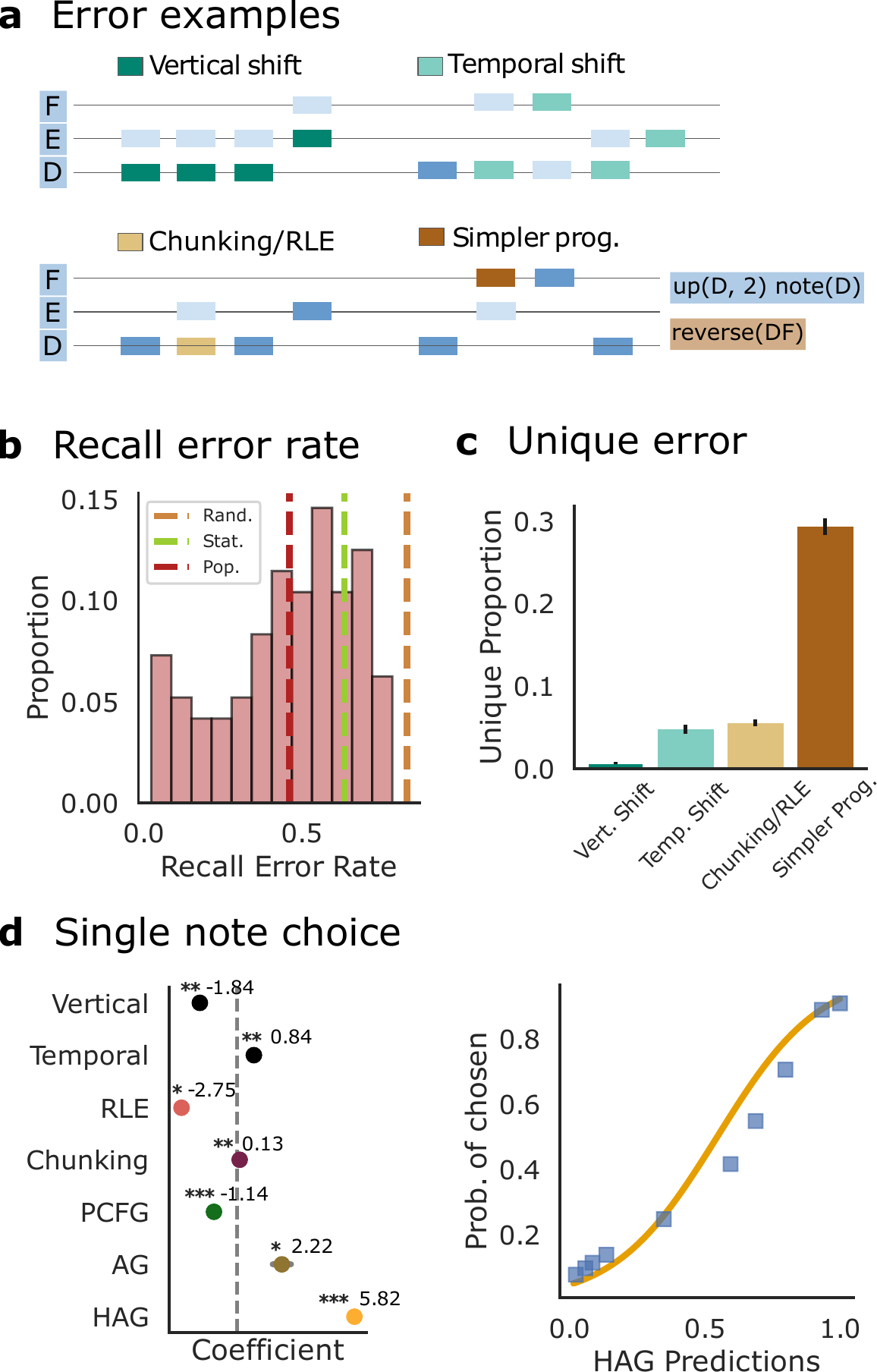}
    \caption{
    \textbf{(a)} Schematic illustration of the four error categories: vertical shifts, temporal shifts, intrusions from earlier notes (RLE) or segments (Chunking), and simplification bias (i.e., corresponds to a simpler program).
    \textbf{(b)} The histogram of error rates across participants in Phase 2. Participants on average (Pop.) perform well above chance (Rand.) and first-order Markovian model (Stat.).
    \textbf{(c)} Frequency of distinct errors across participants and segments (each error counted at most once per category per segments). Program-like errors predominate.
    \textbf{(d)} Logistic-regression coefficients (log-odds) estimating the  notes chosen by participants. The right panel shows marginalized probabilities against HAG-predicted probabilities. Significance stars indicate $^{*}p<.05$, $^{**}p<.01$, $^{***}p<.001$.
    }
    \label{fig:behavior-error}
    \vspace{-1em}
\end{figure}

%% file: figures/behavior/regression/fig-behavior-regression.tex
\begin{figure}[!th]
    \centering
    \includegraphics[width=0.95\linewidth]{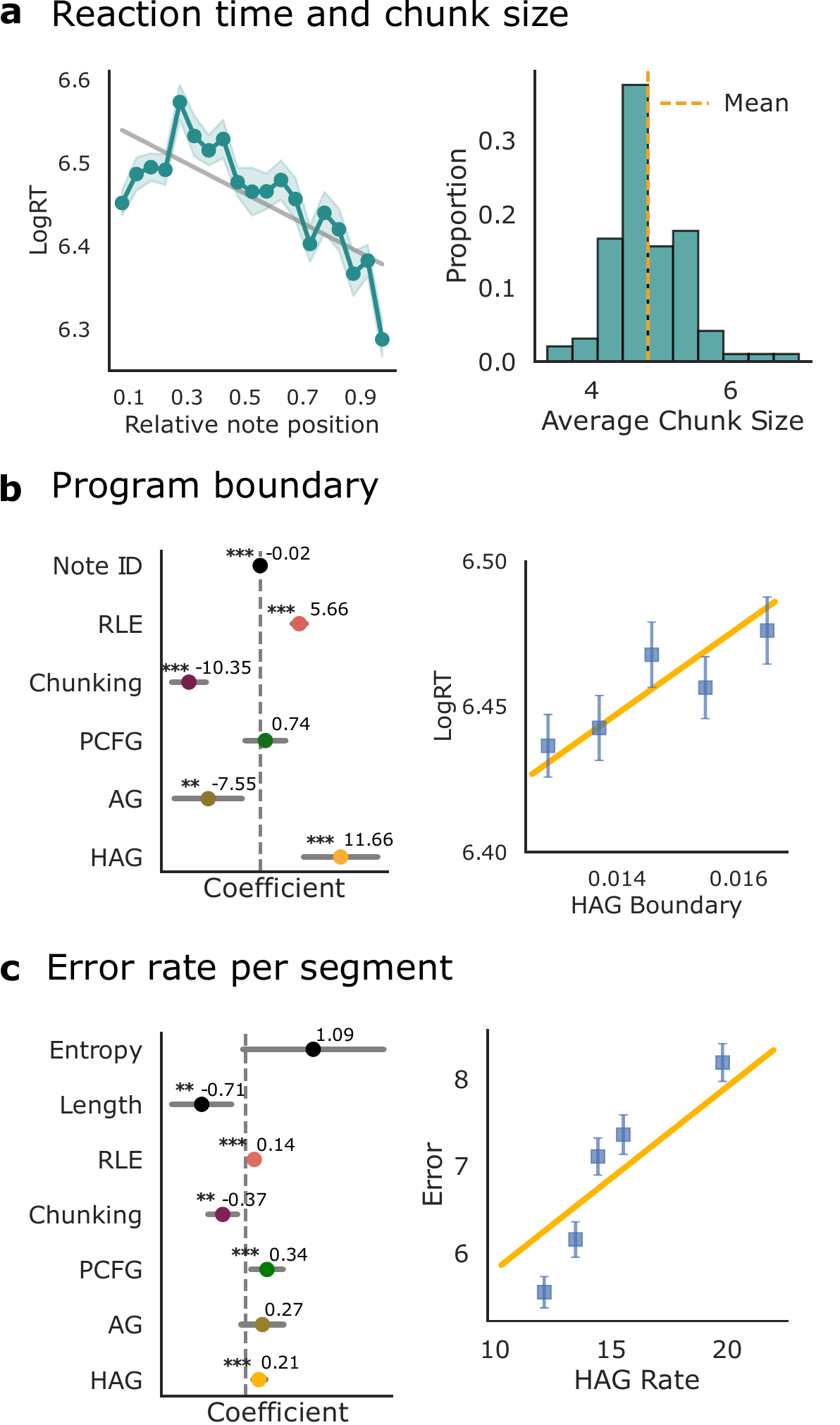}
    \caption{
    \textbf{Program-based predictors explain human recall and reaction-time behavior. }
    \textbf{(a)} Left, mean logRT across relative note position within each segment. Right, distribution of participant-level average chunk size inferred by the reaction time statistics. 
    \textbf{(b)} Program-boundary effects on note-level RT. Left, mixed-effects regression coefficients predicting logRT from model-predicted boundary probability, controlling for note index. Right, mean logRT increased monotonically with HAG-predicted boundary probability. 
    (c) Segment-level recall error as a function of inferred program complexity. Left, mixed-effects regression coefficients predicting segment error from model-based complexity measures. Right, segment error increased monotonically with HAG-inferred rate. 
    Significance stars indicate $^{*}p<.05$, $^{**}p<.01$, $^{***}p<.001$.
    }
    \label{fig:behavior-regression}
    \vspace{-1em}
\end{figure}

%% file: figures/model/fitting/fig-model-fitting.tex
\begin{figure*}[!ht]
    \centering
    \includegraphics[width=0.95\linewidth]{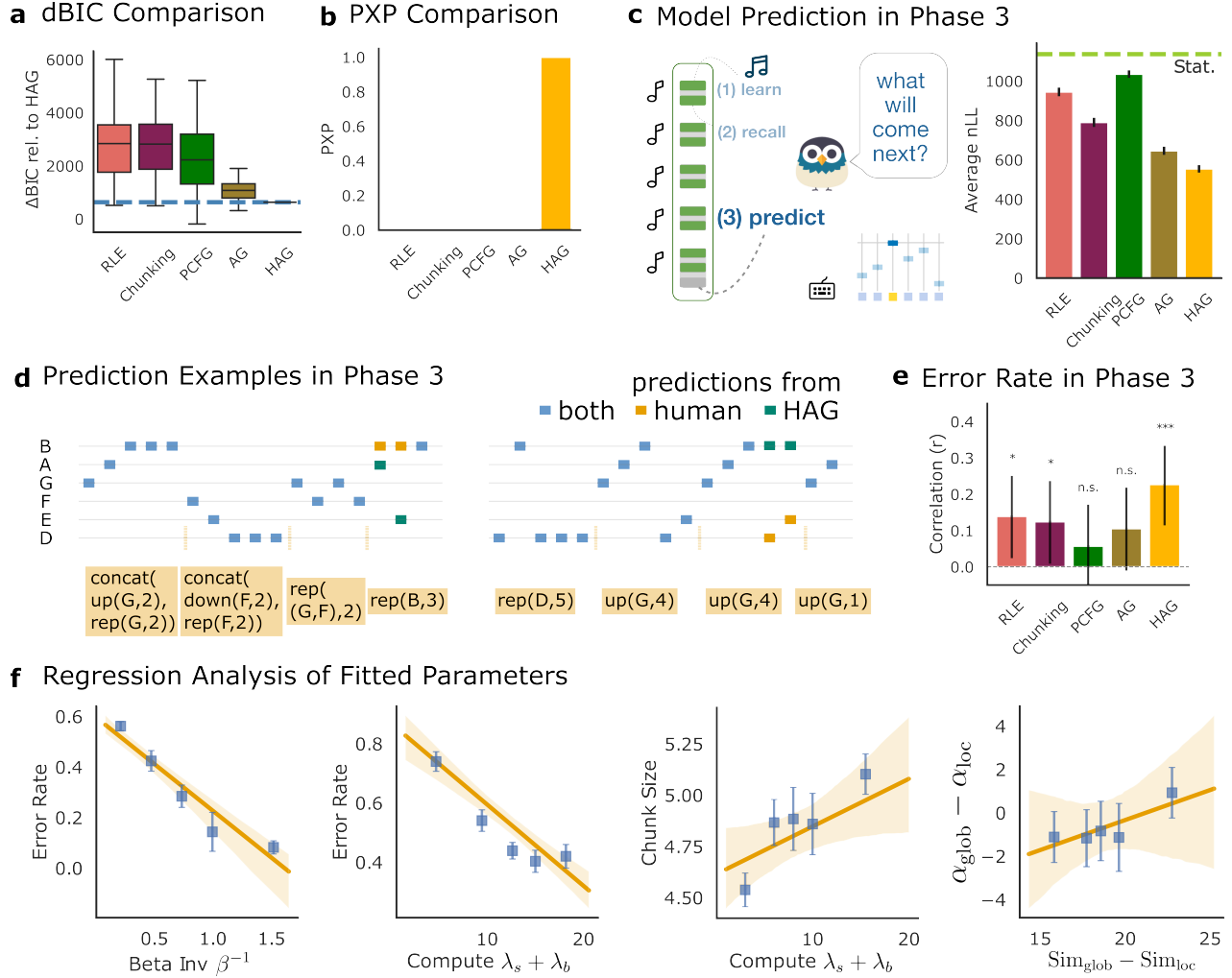}
    \caption{
    \textbf{Model fitting, comparison, and parameter–behavior relationships.}
    \textbf{(a)} Model comparison accounting for complexity, shown as relative $\Delta$BIC values with respect to the best model for each participant. 
    \textbf{(b)} Model comparison using protected exceedance probability (PXP), indicating the likelihood that each model is the most frequent generative model in the population.
    \textbf{(c)} Model-based predictions in the Phase~3. Participants predicted the continuation of a sequence based on what they have seen. Model predictions were generated using the participant-specific fitted model. Bars show the negative log-likelihood of participants’ responses under the model predictions.
    \textbf{(d)} Example predictions produced by two participants (p36 and p71) alongside compositions generated by the HAG model using fitted parameters.
    \textbf{(e)} Correlation between participant Phase~3 error rates and model-predicted error rates, shown separately for each model. Bars indicate Pearson correlation coefficients across participants; error bars denote $95\%$ confidence intervals. Asterisks indicate significance levels.
    \textbf{(f)} Regression analysis relating fitted model parameters to behavioral measures across participants.
    }
    \label{fig:model-fitting}
\end{figure*}

%% file: method.tex
\section*{Methods}

\subsection*{Program induction under resource constraints}
We model human sequence learning as program induction under resource constraints: learners encode sequences into compact programs to reconstruct the original sequence efficiently. To capture a spectrum of possible strategies, we consider models that differ in the expressiveness of their program space and whether they support learning reusable structure. 

\textcolor{rle}{RLE (Run-Length Encoding):}
Programs are built from single operator~$\{\texttt{rep}(n,c)\}$ which encodes repeating note~$n$ for $c$~steps.
After each sequence, the global library~$\lglob$ is updated by adding any $\{\texttt{rep}(n,c)\}$ subprograms selected in the program inferred for that sequence.

\textcolor{chunking}{Chunking:}
Programs are built from a single operator~$\{\texttt{chunk}(n)\}$, which concatenates stored chunks~\cite{wu2023chunking}.
The model assumes that reusable structure strictly takes the form of contiguous subsequences, which are memorized rather than allowing for transformations (compositional or otherwise). 
Similar to RLE, Chunking operates over a single global library~$\lglob$, which can add contiguous subsequences as new chunks whenever doing so decreases the total description length.

\textcolor{pcfg}{PCFG (Probabilistic Context-Free Grammar):}
Programs are built from pre-defined operators: 
\texttt{up}/\texttt{down} (transposition by a step count), \texttt{range} (generate a stepwise sequence from a start note and step size), \texttt{rep} (repeat a note or sequence), \texttt{rev} (reverse order of a sequence), and \texttt{chunk} (verbatim memorization). 
The grammar and its rule probabilities are fixed across sequences, with operators and combinators assigned equal prior probability over well-typed compositions. Each sequence is therefore parsed independently under the same prior. Unlike the library-based models, PCFG has no mechanism for updating its representations across sequences.

\textcolor{ag}{AG (Adaptor Grammar):} 
Programs are built from the same operators as with PCFGs, but AG places a nonparametric prior (adaptor) over reusable programs via a Pitman-Yor process ($\mathrm{PYP}$)~\cite{johnson2006adaptor}. This allows the model to learn which previously inferred programs are worth reusing, rather than treating every sequence as independent. 
Concretely, it defines the distribution~$G \sim \mathrm{PYP}(d,\alpha,H)$ over programs in a global library~$\lglob$, where the base distribution $H$~is given by the PCFG. After each sequence, the posterior over~$G$ is updated, such that programs recurring across multiple sequences become more probable (and therefore cheaper to reuse) when encoding later sequences. RLE and Chunking are special cases of AG, obtained by restricting the operator set to a single repetition primitive \texttt{rep} or to verbatim concatenation \texttt{chunk}, respectively. Both inherit AG's reuse parameters.

\textcolor{hag}{HAG (Hierarchical Adaptor Grammar):}
Programs are built from the same operators as the PCFG and AG.  
However, HAG replaces the single global adaptor in AGs with a \emph{hierarchical} Pitman-Yor process ($\mathrm{HPYP}$) that induces two nested libraries: a sequence-specific \emph{local} library~$\lloc$ and a shared \emph{global} library~$\lglob$. The global library stores reusable programs that have been useful across sequences, whereas the local library stores reusable programs that are especially useful within the current sequence. 
Concretely, the PCFG provides the base structural prior over well-typed combinatory-logic expressions, while the $\mathrm{HPYP}$ determines how inferred programs are reused and updated across the two libraries~\cite{teh2006hierarchical}.

\paragraph{Bayesian program induction with libraries } 
Formally, Bayesian program induction~\cite{goodman2008rational,piantadosi2012bootstrapping,lake2015human,rule2018learning,correa2023exploring} offers a probabilistic realization of the \textit{Language of Thought} hypothesis~\cite{fodor1975language}. 
In our setting, inference is conditioned on hierarchical libraries that bias hypotheses toward previously learned structure. Given an observation $x \in \mathcal{X}$, a candidate program $\pi \in \mathcal{P}$, a global library $\lglob$, and a local library $\lloc$, the posterior can be defined as:
\begin{align}\label{eq:bayes-pi-library}
    p(\pi \mid x, \lglob, \lloc) \propto p(x \mid \pi)\, p(\pi \mid \lglob, \lloc).
\end{align}
The likelihood $p(x \mid \pi)$ measures how well the program reconstructs the observation, whereas the library-conditioned prior $p(\pi \mid \lglob, \lloc)$ favors programs that reuse previously learned structure.

We represent programs in typed combinatory logic (CL)~\cite{schonfinkel_uber_1924,liang2010learning}, where each program is a binary tree. 
For example, a program that captures \textit{ascend by three semitones from note $n$} can be written as $\pi=[\mathbf{CB},[\mathbf{B},\texttt{up},n],3]$. 
Here, combinators are higher-order symbols that specify how arguments are passed between subtrees: $\mathbf{B}$ passes an argument to the right branch, $\mathbf{C}$ passes to the left branch, and $\mathbf{CB}$ passes both branches. 
By contrast, $\texttt{up}$ is a domain-specific transformation operator, and $n$ is an integer-valued argument (Fig.~\ref{fig:main}a).
To ensure that only well-formed programs are considered, we impose type constraints on all symbols and compositions. Base types distinguish semantically different kinds of values, such as notes ($t_n$) and counts ($t_c$), preventing the same numeric token from being reused in incompatible roles. Function types specify valid input-output relations; for instance, $\texttt{up}: t_n \times t_c \to t_n$ maps a note and a step count to a new note.

\paragraph{Two resource constraints: memory and computation.}
Bayesian inference over a compositional hypothesis space of candidate programs is, in principle, unbounded and therefore intractable. 
Because our goal is to model human learners as resource-rational agents rather than unconstrained ideal observers, we impose two resource constraints on program inference (Eq.~\ref{eq:algorithmicRDT-la}).
The first is a \emph{memory} constraint, $R_{\mathrm{mem}}$ (Eq.~\ref{eq:rdt-main-mem}), which limits the complexity of retained programs through their description length. This constraint is implemented as a preference, governed by $\beta$, for shorter and therefore more compressible programs.
The second is a \emph{computational} constraint, $R_{\mathrm{comp}}$ (Eq.~\ref{eq:rdt-main-comp}), which limits the amount of processing the model can perform when proposing candidate programs. 

We formalize the \emph{memory} constraint with rate–distortion theory (RDT) adapted to program induction~\cite{sims2016rate,bates2020efficient,nagy2020optimal,zhou2024harmonizing}.
Let $g:\mathcal{P}\to\mathcal{X}$ denote the interpreter that maps a candidate program to its reconstruction. For a target $x$ and a candidate program $\pi$, execution yields a reconstruction $\hat{x}_\pi = g(\pi)$, and reconstruction quality is measured by a task distortion $d(x, \hat{x}_\pi)$.
Critically, the relevant description length is the complexity of the \emph{representation that is retained}. 
In our setting, the retained object is the reconstruction $\hat{x}$. 
To connect description length to our probabilistic model $p(\pi)$ over programs, we assume programs are encoded with a (near) Shannon-optimal prefix code induced by this prior. This assigns each program $\pi$ a code length~$\ell(\pi) := \lceil-\log p(\pi)\rceil$~\cite{grunwald2008algorithmic,grunwald2007minimum},
which is within an additive constant of~$-\log p(\pi)$ by at most $1$~bit (plus any fixed coding overhead). We then define the description length of a reconstruction $\hat x$ as the length of the shortest code among all programs whose execution yields that reconstruction:
\begin{align}
    L(\hat{x}) := \min_{\pi: g(\pi)=\hat{x}} \ell(\pi) \approx \min_{\pi: g(\pi)=\hat{x}}(-\log p(\pi)).
\end{align}

\noindent The corresponding RD objective is then
\begin{align}\label{eq:algorithmicRDT}
    D(R_{\mathrm{mem}})
    = \inf_{x\in\mathcal X}
        d(x,\hat x) \quad\text{s.t.}\quad
        L(\hat x)\le R_{\mathrm{mem}},
\end{align}
\noindent or equivalently, an optimization over programs while measuring rate at the level of the retained reconstruction.
In practice, we optimize over candidate programs directly and use $-\log p(\pi)$ for the selected program as an approximation to this optimal reconstruction-level description length.

We next model \emph{computation} as a stochastic, resource-bounded search process $A_{\lambda_s,\lambda_b}$ over the program space $\mathcal{P}$. The computational constraint $R_\mathrm{comp}$ is operationalized by two limits: an expected search budget $\lambda_s$ and its realization $N_s \sim \operatorname{Pois}\left(\lambda_s\right) + 1$, which caps the number of candidate programs evaluated, and an expected backtracking budget $\lambda_b$ and its realization $N_b \sim \operatorname{Pois}\left(\lambda_b \right)$, which caps the number of previously selected programs that can be revised in light of new observations.
Let~$q_{\lambda_s,\lambda_b}(\pi\,|\, x)$ denote the resulting distribution over candidate programs, which captures both which programs are reachable and with which probability they are found. The computationally bounded objective is then the expected distortion under this resource-limited search:
\begin{align}\label{eq:boundedRDT}
    D_{\mathrm{comp}}(R_\mathrm{mem};\lambda_s,\lambda_b)
    = 
    \mathbb{E}_{\pi \sim q_{\lambda_s,\lambda_b}(\cdot \,|\, x)}
    \big[ d ( x, \hat{x}_\pi ) \big] \\
    \quad \text{s.t.} \quad
    L(\hat{x}_\pi) \le R_\mathrm{mem} . \nonumber
\end{align}
Since $A_{\lambda_s,\lambda_b}$~is stochastic (i.e., sampling-based), $q_{\lambda_s,\lambda_b}$~explicitly captures discoverability under limited computation.

The corresponding Lagrangian form of Eq.~\ref{eq:boundedRDT} is:
\begin{align}\label{eq:algorithmicRDT-la}
    \mathcal{L}_{\lambda_s,\lambda_b}(x)
    =
    \mathbb{E}_{\pi\sim q_{\lambda_s,\lambda_b}(\cdot \,|\, x)}
    \big[d(x,\hat{x}_\pi)+\beta\,L(\hat{x}_\pi)\big],
\end{align}
\noindent where $\beta$ controls the trade-off between accuracy and complexity, whereas the expected search budget $\lambda_s$ and backtracking budget $\lambda_b$ control exploration and revision during inference, respectively.

\subsection*{Inference over hierarchical libraries. }

The previous section defined program induction for a single sequential pattern under memory and computational constraints. 
We now generalize the formulation to the present setting, in which each data point is an entire melodic sequence rather than a single pattern. For each sequence $X^{(i)}$, the model infers a sequence-level program $\Pi^{(i)}$ built from reusable subprograms $\pi^{(i)}_j$. Each subprogram generates a local reconstruction and is evaluated under the same rate-distortion (RD) trade-off introduced above. Sequence-level inference then operates over the ordered concatenation of these subprograms, determining both which subprograms are reused and how they are assembled to reconstruct the full sequence.

Formally, each observation is a sequence $X^{(i)}=[x^{(i)}_1,\ldots,x^{(i)}_T]\in\mathcal{X}$, explained by a sequence-level program
\begin{align}
    \Pi^{(i)}=\pi^{(i)}_1 \oplus \pi^{(i)}_2 \oplus \cdots \oplus \pi^{(i)}_{M^{(i)}},
\end{align}
where each $\pi^{(i)}_j\in\mathcal{P}$ is a subprogram, $\oplus$ denotes concatenation, and each subprogram generates a contiguous subsequence under the interpreter $g$, such that $\hat{X}^{(i)}_j=g(\pi^{(i)}_j)$. The full reconstruction is therefore
\begin{align}
    \hat{X}^{(i)}=g(\Pi^{(i)})=\hat{X}^{(i)}_1 \oplus \hat{X}^{(i)}_2 \oplus \cdots \oplus \hat{X}^{(i)}_{M^{(i)}}.
\end{align}

\noindent The posterior over sequence-level programs $\Pi^{(i)}$ is
\begin{align}
    p(\Pi^{(i)} \mid X^{(i)}) \propto p(X^{(i)} \mid \Pi^{(i)})\, p(\Pi^{(i)} \mid \lglob,\lloc),
\end{align}
whereas, in practice, the inference is carried out locally over candidate subprograms that explain successive subsequences. 
We next specify the three components of this inference procedure: i) a hierarchical prior over reusable subprograms, ii) a sequence likelihood derived from the subprogram objective, and iii) an approximate online inference algorithm with bounded backtracking.

\paragraph{Hierarchical prior over reusable subprograms.}
The basic unit of reuse in the model is the subprogram $\pi$. We therefore place a prior over subprogram expansions that allows a candidate subprogram to be generated either by composing operators from the underlying typed grammar or by reusing previously inferred subprograms stored in the local and global libraries. 

As the base structural prior, we use a uniform distribution over well-typed CL trees~\cite{liang2010learning}. 
For a subprogram~$\pi=[r,x,y]$ with type~$t$, combinator~$r$ and subtrees~$x,y$, the prior factorizes as
\begin{align}\label{eq:prog-prior}
    \log p(\pi\,|\, t)=\log p(x\,|\, t)+\log p(y\,|\, t)+\log p(r),
\end{align}
\noindent with operators and combinators equiprobable, conditioned on their type.

To model reuse, we place a hierarchical Pitman-Yor process ($\mathrm{HPYP}$) prior over subprogram choices~\cite{teh2006hierarchical}. Let $H$ denote the base distribution over typed subprogram expansions defined by the typed grammar above. 
At the global level, we define a distribution over reusable subprograms, $G_0 \sim \mathrm{PYP}(d_\mathrm{glob},\aglob,H)$, which captures domain-general reuse tendencies across sequences. 
For each sequence~$X^{(i)}$, we then define a sequence-specific distribution over subprograms, $G_i \mid G_0 \sim \mathrm{PYP}(d_\mathrm{loc},\aloc,G_0)$ which biases inference within that sequence toward reusing its own previously inferred subprograms while still permitting reuse of subprograms from the global distribution $G_0$.

We use $G_i$ and $G_0$ to define the predictive probability of the next subprogram. 
As a notational convenience, we distinguish between individual instances of subprograms $\pi$ and subprogram classes $\phi_k$, the latter of which is comprised of identical instances (i.e., $\phi_k$ indexes the $k$-th distinct subprogram observed so far, while $\pi$ refers to a single use). 
For sequence $i$, let $\pi_{<j}^{(i)}$ denote the instances inferred before position $j$. The local library $\lloc$ corresponds to the set of distinct types $\left\{\phi_k\right\}_{k=1}^{K_i}$ assembled from these instances, where $K_i$ is the number of types that have appeared so far in sequence $i$. Let $n_{i k}$ denote how many times type $\phi_k$ has appeared in sequence $i$, and $n_i =\sum_{k=1}^{K_i} n_{i k}$ denote the total number of token assignments in that sequence. Across-sequence quantities $\pi^{(<i)}$, the global library $\lglob$, and the global predictive distribution from $G_0$ are also defined analogously.
The predictive probability of the next subprogram is then
\begin{align}\label{eq:prog-predictive-dist}
    p(\pi \mid \pi_{<j}^{(i)})
    = \underbrace{\sum_{k=1}^{K_i}\frac{n_{ik}-\dloc}{\aloc+n_{i\cdot}}\,\delta_{\phi_k}}_{\text{reuse from local library}}
    + \underbrace{\frac{\aloc + \dloc K_i}{\aloc + n_{i\cdot}}\,G_0}_{\text{reuse global library}}.
\end{align}
The first term assigns mass to each existing local type $\phi_k$ in proportion to how often it has been used so far in the current sequence (with a discount $\dloc$ that flattens the distribution). The second term falls back to the global distribution $G_0$, which itself recursively assigns mass to existing global types or, with residual mass, to a novel subprogram drawn from the base grammar $H$.

The four hyperparameters of the HPYP shape this distribution. The concentration parameters $\aloc$ and $\aglob$ control the rate of innovation at each level: larger values increase the probability of introducing a new subprogram rather than reusing an existing one. When $\aloc>\aglob$, sequence-specific inference is more likely to propose new local subprograms than to draw from globally shared ones. The discount parameters $\dloc$ and $\dglob$ control the diversity of reuse: $d \approx 0$ yields rich-get-richer dynamics in which a few subprograms dominate, whereas $d>0$ flattens the distribution and encourages a heavier-tailed mix of common and rare subprograms.

\paragraph{Tradeoff between distortion and rate as likelihood. }
For a candidate subprogram $\pi^{(i)}_j$ assigned to subsequence $X^{(i)}_j$, we use the Lagrangian form of the RD objective (Eq.~\ref{eq:algorithmicRDT-la}) to define its score:
\begin{align}
    s\!\left(X^{(i)}_j,\pi^{(i)}_j\right)
    =
    -\,d\!\left(X^{(i)}_j,\hat{X}^{(i)}_{j,\pi}\right)
    - \beta\,L\!\left(\hat{X}^{(i)}_{j,\pi}\right),
    \label{eq:rd_score}
\end{align}
where $\hat{X}^{(i)}_{j,\pi}=g(\pi^{(i)}_j)$ is the reconstruction generated by the subprogram, $d(\cdot,\cdot)$ is the Hamming distance between the observed and reconstructed note sequences, and $L(\hat{X}^{(i)}_{j,\pi})$ is the description length of the retained reconstruction. This score favors subprograms that achieve accurate reconstruction with shorter descriptions. Unlike program-induction approaches that impose a hard consistency criterion~\cite{ellis2021dreamcoder,zhao2023model,correa2023exploring}, this formulation allows partially matching hypotheses to remain under consideration, with support decreasing smoothly as distortion increases.

Since new observations are incorporated sequentially, inference must also proceed online. Jointly optimizing over all subprograms in $\Pi^{(i)}$ is computationally intractable for long sequences, so we exploit two properties of the formulation. First, the local score in Eq.~\ref{eq:rd_score} depends only on the subsequence each subprogram explains, so the sequence-level objective decomposes into a sum of local scores (Eq.~\ref{eq:rd_likelihood}).
Second, this additive decomposition lets us search greedily: candidate subprograms are evaluated for each subsequence in turn, committed to one at a time, and revised only over a bounded recent history (the expected backtracking budget $\lambda_b$; see Approximate inference below). The result is an approximation to the joint objective in which earlier commitments influence later ones but cannot be revisited indefinitely, which is also a deliberate consequence of the computational constraint $R_{\text {comp}}$.
Given a sequence-level program $\Pi^{(i)}=\pi^{(i)}_1 \oplus \cdots \oplus \pi^{(i)}_{M^{(i)}}$, we define its log-likelihood by summing subprogram scores:
\begin{align}\label{eq:rd_likelihood}
    \log p\!\left(X^{(i)} \mid \Pi^{(i)}\right)
    \propto
    \sum_{j=1}^{M^{(i)}}
    s\!\left(X^{(i)}_j,\pi^{(i)}_j\right).
\end{align}

\paragraph{Approximate inference. }
Exact inference over the joint posterior
$$p\left({\Pi^{(i)}}{i=1}^N,{G_i}{i=1}^N, G_0 \mid {X^{(i)}}_{i=1}^N\right)$$
is intractable. We approximate it with a resource-bounded procedure that processes sequences in order and, within each sequence, processes subsequences from the beginning to the end(Algorithm~\ref{alg:inference}). Three operations are executed at each step.

(1) \textit{Within-sequence search}.
For each sequence~$i$, inference proceeds over subsequences. At each step, we perform Gibbs sampling under a fixed search budget~$N_s \sim \operatorname{Pois}\left(\lambda_s\right) + 1$. Candidate subprograms are proposed from the $\mathrm{HPYP}$~mixture, either by reusing from the local and global libraries, or by generating new candidates from the base structure prior (Eq.~\ref{eq:prog-prior}), which ensures type-consistent syntactic validity. Each candidate~$\pi$ is evaluated using the loss~$\mathcal{L}_{\lambda_s,\lambda_b}(x)$ in Eq.~\ref{eq:algorithmicRDT-la} such that subprograms with lower loss are more likely to be selected, although higher-loss candidates may still be chosen with non-zero probability.

(2) \textit{Depth-limited backtracking}.
To allow for local revisions when seeing new data, the inference process maintains a stack of accepted subprograms and allows up to $N_b \sim \operatorname{Pois}\left(\lambda_b\right)$~backtracking steps. When no local revision improves compression score, the most recent decision is used. This is similar in spirit to history-limited forward induction, analogous to bounded dynamic programming. The realized $N_b=0$ means a purely greedy strategy, while increasing~$N_b$ progressively approximates optimal dynamic-programming solutions.

(3) \textit{Library ($\mathrm{HPYP}$) updates}.
After subprogram assignments $\pi^{(i)}_j$ for sequence~$X^{(i)}$ are selected during inference, we update the hierarchical reuse prior (Eq.~\ref{eq:prog-prior}). In the local library, each selected subprogram instance $\pi^{(i)}_j$ is matched to a distinct reusable subprogram class $\phi_k$ and increments its count $n_{ik}$. At the global level, the corresponding cross-sequence counts are updated in $G_0$ with a complete program is selected $\Pi^{(i)}$. These count updates modify the predictive distributions $G_i$ and $G_0$, increasing the probability of reusing subprograms that proved useful in previous inferences.

The three operations are executed in a single pass over sequences: each sequence is processed once, in the order it is encountered. Within a sequence, recent subprogram commitments may be revised via the backtracking budget $N_b$ as new observations arrive; but once a sequence's processing concludes, its subprogram assignments and the resulting library updates are fixed, and later sequences do not trigger reanalysis of earlier ones. This single-pass schedule across sequences mirrors the path-dependent nature of human learning---what a learner takes from sequence $i$ depends on the library inherited from sequences $1, \ldots, i-1$. 

\subsection*{Dataset and simulations}
\subsubsection*{Melodic sequence dataset}
We converted a dataset of real-world MIDI files~\cite{garciavalencia2020sequence} into structured piano-roll sequences using the following pipeline. 
1) \emph{Instrument filtering:} we retained only files whose primary instruments are piano, brass, reed, or synth lead in order to retain melodic as opposed to ambient (e.g., synth pad) or rhythmic samples (e.g., drums).  
2) \emph{Pitch normalization:} we mapped notes to a single octave comprising 12 pitch classes, $\{C, C^\sharp, D, D^\sharp, E, F, F^\sharp, G, G^\sharp, A, A^\sharp, B\}$. 
3) \emph{Note constraint:} we only included in our stimulus set sequences containing exactly six unique pitch classes, such that participants could comfortably execute the sequences with both hands on their keyboard during the task. 
4) \emph{De-duplication:} we removed duplicate note sequences that resulted from pitch normalization. 
5) \emph{Length constraint:} we retained sequences with a total length between $80$ and $120$ notes to ensure sufficient material for segmentation, while avoiding extreme memory load.
We then followed the dataset’s original training–evaluation split~\cite{garciavalencia2020sequence}.
After preprocessing, we were left with $298$~sequences for training and $96$~sequences for testing. 

\subsubsection*{Simulations} 
We randomly sampled $50$~melodic sequences from the training set. In all simulations, we treated the model as a plausible observer with access to complete sequences. Each simulation was repeated with $100$~random seeds, which varied both the presentation order and the stochastic choices involved in program selection. Results were averaged across seeds.

\subsection*{Experiment}
\subsubsection*{Participants and design}
We recruited $N=100$~participants ($56$~female; $M_{\text{age}}=34.28$, $SD=11.23$) via Prolific. Informed consent was obtained from all participants prior to participation and the study was approved by the Ethics in Psychological Research Commission of the University of Tübingen (Wu\_2021/0124/213).
We excluded $4$~participants because of technical issues with data saving. No participants were excluded based on task performance or a priori criteria.
Participants received a base payment of \pounds{}$4.00$ plus a performance-contingent bonus (\pounds{}$2.68 \pm 1.17$). The task lasted approximately~$46.87 \pm 28.54$\,min.

\subsubsection*{Materials and procedure}
The experiment was a sequence-learning task (Fig.~\ref{fig:main}) in which participants \emph{learned}, \emph{recalled}, and \emph{continued} melodic sequences. 
Participants completed the study online in a browser environment. After providing informed consent, they received on-screen instructions and completed an interactive tutorial. Comprehension was verified with three multiple-choice questions, which participants had to correctly answer before proceeding. 

At the start of the experiment, each participant was randomly assigned $5$~melodic sequences from the full set of training sequences to ensure a reasonable experiment length, and to avoid both fatigue and sequence-specific biases. Each sequence was subject to a random transposition to prevent over-representation of specific pitch values, while preserving the relative interval structure. An overview of the sequence statistics is provided in Figure~\ref{fig:si:melody-stats}. 
We index the five sequences by $i \in \{1,\ldots,5\}$. Each sequence $X^{(i)}=\left[x_1^{(i)}, \ldots, x_{T^{(i)}}^{(i)}\right]$ contained a string of discrete MIDI notes. After preprocessing (see above), each note took one of six values: $x_t^{(i)} \in\{1, \ldots, 6\}$.

Additionally, each sequence was further divided into six segments, denoted $X^{(i,1)},\ldots,X^{(i,6)}$, using kernel change-point detection~\cite{celisse2017new} as implemented in the \textit{ruptures} Python package. 
This reduced the memory demands of the task by presenting long sequences (usually over $100$ notes) in smaller, more manageable units ($12-16$ notes), while still preserving their higher-order sequential structure. 
This produced segments of comparable complexity: mean segment length was $14.89 \pm 0.41$ notes, first-order transition entropy was $1.21 \pm 0.04$ bits/note, and repetition complexity was $21.23 \pm 0.87$ symbols (see Fig.~\ref{fig:si:melody-stats} for full distributions).

Segments $X^{(i,1)},\ldots,X^{(i,5)}$ constituted the training portion of each sequence and were presented to participants in alternating learning and recall phases. In the learning phase, participants observed each segment note-by-note, and were required to press the corresponding key for each visualized note before the next note was shown. In the subsequent recall phase, they sequentially reproduced the same segment from memory note-by-note, but without any visual cues. Participant's responses during the recall for segment $j$ of sequence $i$ are denoted  as $\tilde{X}^{(i,j)}$. Reaction times of each keystroke are denoted $\textrm{RT}^{(i,j)}$, with analyses conducted on $\mathrm{LogRT}$.

The sixth segment, $X^{(i,6)}$, served as a held-out ground truth for the Phase 3 continuation task and was never presented during learning or recall. Instead, after recalling segment $X^{(i,5)}$, participants proceeded directly to a continuation phase, in which they were asked to improvise a completion consistent with the structure of the preceding five segments. This unseen sixth segment therefore provided a ground-truth continuation for evaluation.

Throughout the experiment, participants were incentivized to maximize accuracy in all phases. Participants earned a performance-based bonus computed from note-level accuracy (binary) in all three phases. 
A brief post-task questionnaire collected demographic information and self-reported strategies. 
The performance summaries for each stage are shown in Figure~\ref{fig:si:error-rates-and-reaction-time}.

\subsection*{Behavior analyses}

\subsubsection*{Error patterns}
We analyzed recall errors by assigning each error to one of four categories (Fig.~\ref{fig:behavior-error}a).
For each sequence and participant, we scanned all contiguous windows $[a\!:\!b]$ and evaluated candidate error categories whenever at least one note differed, i.e., $d(X_{a:b},\tilde{X}_{a:b})\ge 1$. 

\noindent\textbf{Vertical shifts} capture errors where participants recalled the correct relative interval structure, but transposed the true note up or down in pitch. We computed element-wise pitch differences~$d_j=\tilde{x}_j-x_j$. If there was a constant pitch offset across the window $d_a=\cdots=d_b=c\neq 0$, we classified the error as a vertical shift with magnitude~$|c|$.

\noindent\textbf{Temporal shifts} correspond to when participants recalled the correct notes, but at the wrong temporal position. We tested whether the recalled notes matched the ground truth after a uniform index shift in time. Specifically, if there exists a lag~$s$ such that~$a' = a+s \ge 0$, $b' = b+s \le T$, and~$\tilde{X}_{a':b'} = X_{a:b}$, then the window was classified as a temporal shift with lag~$s$.

\noindent\textbf{Repetition errors} capture when participants produced notes that matched earlier patterns rather than the current target. We operationalized two forms of repetition errors within a recalled window~$\tilde{X}_{a:b}$:
\emph{(i) Run repetition} corresponds to repeating (once or multiple times) the immediately preceding recalled note. Let~$p=\tilde{X}_{a-1}$ (with~$a>0$) and let $\mathbf{1}$~denote an all-ones vector of length~$W=b-a+1$. If~$\tilde{X}_{a:b}=p\,\mathbf{1}$, we labeled this as a run repetition and report the run length~$W$;
\emph{(ii) Chunk repetition} corresponds to repeating (once or multiple times) an earlier sequence of multiple notes. Let~$j$ index the current segment. For each earlier segment~$j' < j$ within the same melody, we checked whether the recalled window copied a contiguous earlier ground-truth sequence, i.e., there exist indices~$u\!:\!v$ within~$X_{j'}$ such that~$\tilde{X}_{a:b}=X_{u:v}$ while~$X_{a:b}\neq X_{u:v}$.
If so, we labeled the window as a chunk repetition with segment lag~$\Delta j = j-j'$.

\noindent\textbf{Simpler programs} capture errors in which an incorrect participant response was better explained by a more compressible program than the ground-truth pattern. 
To make this comparison model-agnostic, we used the same operators that all program-based models share---and exhaustively enumerated all type-consistent programs up to tree depth $D=2$, generating a candidate program set. This candidate set was fixed in advance and does not depend on any model's learned library, so the categorization does not advantage any of the models we later compare.
For each window $[a: b]$, we scored both the ground-truth subsequence $X_{a: b}$ and the participant's response $\tilde{X}_{a: b}$ under the same candidate program set. We score them by 
\begin{align}
    v(X)=\log \left(1-\frac{d(X,g(\pi))}{|W|}+\varepsilon\right)+\frac{\log p(\pi)}{|W|},
\end{align}
where $|W|$ is the window length, $d(X, g(\pi))$ is the Hamming distortion, $\log p(\pi)$ is the program's code length under the uniform prior, and $\varepsilon=10^{-3}$. The two terms reward reconstruction accuracy and penalize program complexity, both normalized by $|W|$ so that scores are comparable across windows.
A window was labeled a simpler-program error if $v\left(\tilde{X}_{a: b}\right)-v\left(X_{a: b}\right)>0$ i.e., some program compressed the response better than any program compressed the ground truth---and report the advantage $v\left(\tilde{X}_{a: b}\right)-v\left(X_{a: b}\right)$.

\subsection*{Plausible observers}
To generate trial-by-trial predictors for the behavioral analyses, we applied a family of resource-bounded \emph{plausible observers}. We use the term \emph{plausible observer} rather than \emph{ideal observer} because these models are explicitly constrained by limited memory and computation, and therefore do not assume exhaustive search over all possible programs or unlimited representational resources.
These observers were not fit to participant data, thus potentially advantaging more complex models. Instead, we used them to derive theory-driven predictions by simulating across a range of resource settings, asking whether the qualitative predictions of the resource-rational sequence-learning framework remain stable across different levels of memory and computational constraint, rather than depending on a narrowly tuned parameterization. 

To do so, we evaluated the models on a discrete grid of parameter values chosen to capture a broad range of memory and computational resources, which also matched closely to the empirical distribution of participant parameter estimates (Fig.~\ref{fig:si:fitted-param-distribution}), albeit with participants having a narrower range of resource budgets. This allowed us to test whether the qualitative predictions were confined to the human-like range or remained robust under slightly less stringent constraints.
Concretely, we varied memory limit~$\beta=[0.1,0.5,1.0,5.0,10.0]$ and the computational limit $\lambda_s \in [1,5,10,20]$ and $\lambda_b  \in [0,1,3,5,10]$. 
For the realizations $N_s$ and $N_b$, the grid values denote the means of Poisson distributions rather than fixed deterministic budgets ($N_s \sim \operatorname{Pois}\left(\lambda_s\right) + 1$ and~$N_b \sim \operatorname{Pois}\left(\lambda_b\right)$). We used this formulation because search proposals and backtracking operations are count-valued computational resources, and the bounded inference procedure is stochastic: the effective amount of search and revision can vary across sequences even for the same observer. Thus, $\lambda_s$ and $\lambda_b$ capture expected computational resources, while the realized budgets on a given run are sampled from Poisson distributions with those means.
For the models involved libraries, we varied the global and local library-construction costs jointly, setting $\aglob=\aloc \in [1.0,5.0,10.0,15.0]$. These ranges were selected to cover both relatively strict and relatively lax constraints, with step sizes chosen to balance computational tractability against coverage of the empirically relevant parameter regime.

We implemented the five candidate models described above---\textcolor{rle}{RLE}, \textcolor{chunking}{Chunking}, \textcolor{pcfg}{PCFG}, \textcolor{ag}{AG}, and \textcolor{hag}{HAG}---to generate competing predictions for behavior.
For every sequence, each observer returns a candidate program together with its associated rate, distortion, and reconstructed sequence; these quantities are then used as regressors in the behavioral models. All observers have the same setting as in the simulation section.

\subsubsection*{Predicting choices}
To characterize how different models predicted participants' recall response in Phase~2, we fit a hierarchical mixed-effects multinomial regression model (Fig.~\ref{fig:behavior-regression}). 
The outcome variable was the selected note on each response, encoded as a categorical variable with six possible pitches. 
two descriptive control models for non-structured errors (vertical shift and temporal shift), and five plausible-observer models for structured errors. Specifically, repetition-like errors were captured by \textcolor{rle}{RLE} and \textcolor{chunking}{Chunking}, whereas errors favoring a more compressible latent program were captured by \textcolor{pcfg}{PCFG}, \textcolor{ag}{AG}, and \textcolor{hag}{HAG}. 
For each observation, predictor values were aligned over a common ordering of the six candidate notes, resulting in a tensor $X$ of shape ($N \times C \times M$), where $N$~is the number of observations, $C=6$~the number of possible notes, and $M=7$~the number of models.

Each predictor set $m$ was assigned a population-level weight $\omega_m \sim \mathcal{N}(0,\sigma_\omega^2)$, capturing its average contribution to note choice, together with participant-specific random slopes $u_{p,m} \sim \mathcal{N}(0,\sigma_{u,m}^2)$ allowing for their contributions to vary across participants. We also included participant-specific note intercepts $v_{p,c}$, constrained to sum to zero across notes within each participant, to capture baseline note preferences.

The linear predictor combined fixed and random effects as follows:
\begin{align}
    \eta_{n,c} = \sum_{m=1}^M X_{n,c,m}\,(\omega_m + u_{p[n],m}) + v_{p[n],c},
\end{align}
where $p[n]$ denotes the participant associated with observation $n$. Choice probabilities were then given by a softmax function: 
\begin{align}
    p_{n,c} = \frac{\exp(\eta_{n,c})}{\sum_{c'} \exp(\eta_{n,c'})},
\end{align}
and participant responses were modeled as categorical draws from this distribution. Inference was performed using automatic differentiation variational inference (ADVI~\cite{kucukelbir2017automatic}) to approximate the posterior distribution, followed by posterior sampling from the variational approximation. This approach was chosen for computational tractability given the large number of observations and hierarchical parameters. 
Results reported in Figure~\ref{fig:behavior-regression} reflect posterior summaries of fixed and random effects.

\subsubsection*{Reaction time analyses }
To infer whether participants ``chunk'' notes together during recall, we detected sudden positive ``jumps'' in consecutive $\mathrm{LogRT}$ within each recalled segment with~$\Delta r_j=\log r_{j+1}-\log r_j$ for~$j\in [1, \ldots, L-1]$. This corresponds to a slowdown in reaction, following a window of relatively faster responses.
A position $j$ is marked as a chunk boundary if~$\Delta r_j>\mu_{\Delta r}+ \sigma_{\Delta r}$ where~$\mu_{\Delta r}$ and~$\sigma_{\Delta r}$ are the mean and standard devitation of~$\left\{\Delta r_j\right\}$ within the same segment.
Since these boundaries partition the segment into subsequences, we define chunk size as the length of these consecutive notes (subsequence length). If no jump is detected, the whole segment is treated as one chunk.

We first characterize baseline temporal dynamics in sequence processing (i.e., faster RTs over trials) by fitting a linear mixed-effects model predicting log-transformed RTs from the serial note position within a segment $\mathrm{LogRT} \sim \mathrm{NoteID} + (1 \mid \mathrm{ParticipantID})$. 
Here, $\mathrm{NoteID}$ represents the ordinal position of each note within its segment, and random intercepts account for individual differences in baseline response speed. We used restricted maximum likelihood (REML) estimation to fit the model (Fig.~\ref{fig:si:learning-dynamics}c).

To test whether human processing time reflects program-based compression (Fig.~\ref{fig:behavior-regression}c), we fitted a mixed-effects model incorporating hierarchical program boundary predictions from our set of plausible observers alongside baseline comparisons:
$\mathrm{LogRT} \sim \mathrm{SequenceID} + \mathrm{SegmentID} + \mathrm{NoteID} + \mathrm{RLE} + \mathrm{Chunking} + \mathrm{PCFG} + \mathrm{AG} + \mathrm{HAG} +(1 \,|\, \mathrm{ParticipantID} )$
where $\mathrm{SequenceID}$ and $\mathrm{SegmentID}$ capture hierarchical positional effects, while boundary predictors represent normalized segmentation strengths. Positive regressors thus indicate predictive power in explaining jumps in RTs.

\subsubsection*{Distortion analyses }
To test whether recall errors reflect only surface-level sequence statistics or instead track the representational costs implied by program compression, we analyzed segment-level error rates as a function of both stimulus properties and plausible-observer program complexity (Fig.~\ref{fig:behavior-regression}c). 
For each participant, we summarized each training segment and computed objective descriptors of its local statistics, then asked whether model-inferred program rate explained variance in errors beyond these descriptors.
The sequence descriptors below serve as stimulus-level covariates (i.e., nuisance variables) capturing generic difficulty, such as longer sequences or entropy. Including them allows us to isolate the unique relationship between errors and the model-derived program rate (description length) for each model. 

For each segment~$X^{(i,j)}$, we computed the following quantities:
(i) Length~$T$ measures number of notes in~$X^{(i,j)}$; 
(ii) Transition entropy~$H_1(X)$ measures the local unpredictability of the next note given the current one, treating each sequence as a first-order Markov chain over the note alphabet $\mathcal{A}$:$H_1(X) = -\sum_{n \in \mathcal{A}} p(n) \sum_{m \in \mathcal{A}}p(m \mid n) \log_2 p(m \mid n) \quad \text{(bits/note)}.$

Together with the complexity (i.e., rate) outputs of our plausible observers, the full mixed-effects model predicted participants’ error rates using:~$\mathrm{ErrorRate} \sim \mathrm{SequenceID} + \mathrm{SegmentID} + T + H_1  + \mathrm{RLE} + \mathrm{Chunking} + \mathrm{PCFG} +\mathrm{AG} + \mathrm{HAG} + (1 \,|\, + \mathrm{ParticipantID})$. 
Intuitively, a positive relationship indicates that higher human error rates are correspond to higher complexity in a given a plausible observer, while accounting for lower-level factors (i.e., length and entropy).

\subsection*{Model-based analyses}
\paragraph{Model fitting. }
We estimated five free parameters for the full HAG model, with other models having a subset as nested variants. The parameters were defined as follows:

1) Memory budget~$(\beta)$: This parameter~$(\beta \geq 0)$ governs the tradeoff between minimizing distortion and complexity in the objective function~$D+\beta R$ (Eq.~\ref{eq:rdt-main-mem}). A higher~$\beta$ indicates a stronger preference for concise (shorter) programs. Values were optimized via grid search in log-space:~$\log (\beta) \in\{0,0.5, \ldots, 3.5\}$. Values outside this range produced degenerate behavior (either verbatim memorization or near-empty programs).

2) Computational budget~$\left(\lambda_s, \lambda_b\right)$: These parameters constrain the effort in proposing program candidates. To model cognitive variability, we assumed the realized computational effort for proposing ($N_s$) and revising ($N_b$) programs follows a Poisson process. For each participant, segment, and sequence, budgets were sampled as~$N_s \sim \operatorname{Pois}\left(\lambda_s\right) + 1$ and~$N_b \sim \operatorname{Pois}\left(\lambda_b\right)$. The rate parameters were defined along discrete sets~$\lambda_s \in\{1,3,5,7,9\}$ and~$\lambda_b \in\{0,1,2,3,4\}$, where higher rates reflect greater computational resources. The grid for $\lambda_s$ starts at 1 because at least one candidate must be proposed for inference to proceed, whereas $\lambda_b$ can take value 0, corresponding to purely greedy (non-revising) inference. The upper bounds were set by computational tractability, i.e., higher values made the simulation procedure infeasible within our compute budget (around 3 days per run).

3) Reuse parameters~$\left(\aglob, \aloc\right)$ control the balance between innovation (generating new programs) and reuse. $\aglob$~governs the global library, while $\aloc$~governs the local library. Higher values favor the generation of new programs over the reuse of existing cached operators. These are optimized via a grid search in log-space:~$\log \aloc \in \{-0.5,0,0.5,\ldots,2.5\}$.

The other four models include a subset of these parameters. The simplest PCFG model include only resource-related parameters ($\beta, \lambda_s, \lambda_b$). The library-based Chunking, RLE, and AG models introduce a global library governed by~$\aglob$. In the limit where~$\aglob$ is high (minimizing reuse), AG converges to PCFG. 
Finally, the HAG extends AG by introducing a local library governed by~$\aloc$. When $\aloc$~is high (suppressing local reuse), HAG reduces to AG.

To ensure robust inference, each model was fitted using $50$~independent random seeds. For each fitted model, we drew $50$~posterior samples of the induced programs to construct a categorical distribution over note space. Model log-likelihoods were computed relative to this distribution.

\paragraph{Simulation-based evaluation of fitted models.}
To complement the trial-by-trial fits, we evaluated each fitted model by simulating behavior from participant-specific posterior samples and comparing the resulting predictions with observed behavior in Phases 2 (recall) and 3 (prediction). For each participant and candidate model, we used the best-fitting parameter estimates from the trial-by-trial fitting procedure, together with the inferred local and global libraries where applicable. To capture uncertainty arising from stochastic inference, we retained 50 posterior samples obtained from independent runs initialized with different random seeds. Each sample thus specified a complete fitted model for that participant, including both parameter values and inferred representational structure.

\emph{Phase 2 (recall).} We applied the fitted samples to simulate recall for each subsequence presented during learning, extracting two statistics: simulated error rate (Hamming distance between simulated and target sequences, averaged across samples) and simulated chunk size (mean length of subprograms in the inferred representation). We related these simulated quantities to observed behavior in two ways. In Figure~\ref{fig:model-fitting}f, we relate observed behavior directly to participants' fitted parameters (i.e., without the intervening simulation step), to show how individual differences in resource budgets and reuse parameters track individual differences in error rate and chunk size. In Figure~\ref{fig:si:simulation-learner}, we relate observed behavior to the simulated quantities themselves — a stronger test, because it asks whether the generative model under each participant's fitted parameters produces participant-like behavior, not only whether the fitted parameters happen to be correlated with that behavior. For the latter, we fit mixed-effects regressions predicting the observed measure jointly from the simulated measures of all candidate models, with participant as a random effect, and plot the partial fixed effect of each focal model's simulated predictor while holding the others constant.

\emph{Phase 3 (continuation).} We used the fitted models and libraries as generative models of sequence continuation. For each participant, we generated synthetic continuations from the 50 posterior samples and aggregated them into a predictive categorical distribution over the six possible notes at each continuation position. We then computed the log-likelihood of the participant's actual continuation responses under this predictive distribution.

%% file: si.tex
\clearpage
\begin{center}
    \LARGE \textbf{Supplementary Information for} \\[1.5em]
    \large Path-dependent program induction under resource constraints explains human sequence learning \\[1em]
    \large 
    Hanqi Zhou\textsuperscript{1,2,3,4,*}, David G. Nagy\textsuperscript{2,3,4}, Peter Dayan\textsuperscript{1,4}, Charley M. Wu\textsuperscript{2,3,4} \\[1em]
    \textsuperscript{1}University of Tübingen, Tübingen, Germany \\
    \textsuperscript{2}Center for Cognitive Science, Technical University Darmstadt, Darmstadt, Germany \\
    \textsuperscript{3}Hessian.AI, Darmstadt, Germany \\
    \textsuperscript{4}Department of Computational Neuroscience, Max Planck Institute for Biological Cybernetics, Tübingen, Germany \\
    \vspace{1em}
    *hanqi.zhou@uni-tuebingen.de \\

\end{center}

\section*{Supplemental Simulation Methods}

\subsection*{Inference}

We describe the resource-bounded inference procedure used to approximate the posterior over subprograms and hierarchical libraries. Because exact inference over long sequences is intractable under our typed program grammar, we adopt a streaming, piecewise search that maintains a stack of accepted subprograms and revises only a bounded history. 
For each subsequence, the algorithm generates well-typed candidate subprograms using the $\mathrm{HPYP}$~predictive mixture (local reuse, global reuse, or innovation from the base grammar; Eq.~\ref{eq:prog-predictive-dist}), scores each candidate under the rate-distortion (RD) objective (Eq.~\ref{eq:algorithmicRDT-la}), and then stochastically selects a subprogram. 
To mitigate myopic commitments, it also performs backtracking by re-optimizing the recent subprograms when this improves total compression. Algorithm~\ref{alg:inference} gives the full pseudocode.

\input{figures/simulation/rd-simulation/algorithm}

\subsection*{Simulation parameters}

We evaluated all models across a broad grid of parameter settings intended to 
(i) span qualitatively distinct resource regimes (weak to strong constraints), (ii) include the resource levels that best fit human behavior while also stress-testing the models outside that regime, 
and (iii) ensure conclusions do not depend on a narrow tuning of computational or memory assumptions. 
Because these parameters govern resource limitations during learning, sweeping across them allows us to test whether qualitative model comparisons (e.g., HAG’s advantage) are robust across plausible cognitive capacities.
\begin{itemize}
    \item Memory-rate parameter~$\beta$: Controls the trade-off between simplicity and memory capacity. Larger values favor shorter subprograms, reflecting stronger constraints on memory and a preference for parsimony. $\beta$~was varied across~$\{0.1, 0.5, 1.0, 5.0, 10.0\}$
    \item Expected search budget~$\lambda_s$: Controls the expected number of subprogram proposals evaluated at each time step. Higher values mean greater computational capacity, as more candidates are considered before selecting one. We varied this budget~$\lambda_s$ across~$\{1, 5, 10, 20\}$; realized values~$N_s \sim \operatorname{Pois}\left(\lambda_s\right)+1$ were sampled per run from a Poisson distribution with the corresponding mean. 
    \item Expected backtracking budget~$\lambda_b$: Controls the extent of backtracking within each sequence. Larger values allow the model to revisit and revise previously chosen subprograms more frequently, again reflecting greater computational capacity. We varied this budget~$\lambda_b$ across~$\{0, 1, 3, 5, 10\}$; realized values~$N_b \sim \operatorname{Pois}\left(\lambda_b\right)$ were sampled per run from a Poisson distribution with the corresponding mean.
    \item reuse parameters in the Pitman–Yor process~$\aloc, \dloc, \aglob, \dglob$: Control the tendency to reuse existing subprograms in either local or global libraries. Higher values of~$\aloc, \aglob$ correspond to weaker reuse tendencies, indicating increased computational capacity. We varied~$\aloc$ and~$\aglob$ across~$\{1.0, 5.0, 10.0, 15.0\}$, while holding~$\dloc$ and~$\dglob$ constant at~$0.2$ to avoid confounding effects~\cite{teh2006hierarchical}.
\end{itemize}

\subsection*{Supplemental simulation results}

\input{figures/simulation/si-eval/fig-eval-rd-curve}

\paragraph{Generalization performance on unseen sequences.}
In Figure~\ref{fig:rd-simulation}, HAG achieved the most favorable RD trade-off with AG typically second-best.
A paired-samples Wilcoxon test confirms that HAG attains significantly lower loss (distortion and rate) than AG ($z(99)=1183, p<.001, r=-.531, \text{CI}=[-.702,-.337]$). 
AG in turn outperformed RLE ($z(99)=1387, p<.001, r=-.451, \text{CI}=[-.649, -.241]$). 

In Figure~\ref{fig:si:eval-rd-curve}, we report generalization performance under the parameter setting~$\lambda_s=10, \lambda_b=1, \aglob=\aloc=1.0$, and~$\dglob=\dloc=0.2$, chosen to lie near the central tendency of the parameters range. (HAG vs. AG:~$z(99)=81, p<.001, r=-.968, \text{CI}=[-.991, -.925]$; AG vs. RLE:~$z(99)=4008, p=.999, r=.587, \text{CI}=[.408, .740]$). After training the library on $50$~sequences, we evaluated compression performance on a separate set of previously unseen $50$~sequences. 
During evaluation, we reused the trained library but varied the number of optimization iterations. 

Increasing the number of iterations improved absolute compression performance but did not alter the rank ordering of models. HAG consistently outperformed AG across all iteration counts. However, varying $\beta$ did affect relative performance: when the memory constraint is made sufficiently strong (e.g., $\beta=10.0$), the objective becomes dominated by rate, reducing the advantage of maintaining hierarchical libraries since all models are pushed toward maximally compact representations regardless of structure, and HAG's advantage over AG diminishes (when $\beta=10.0$, HAG vs. AG:~$z(99)=396, p=.010, r=-.38, \text{CI}=[-.66, -.08]$). 

\input{figures/simulation/si-rd-over-time/fig-rd-over-time}
\paragraph{RD performance as training data increases.}
Figure~\ref{fig:si:rd-over-time} shows how library-based models improve compression efficiency as they are exposed to more training sequences, with darker lines corresponding to more training data. 
While \textcolor{pcfg}{PCFG} shows no change in its RD curve, all models equipped with library learning (\textcolor{chunking}{Chunking}, \textcolor{rle}{RLE}, \textcolor{ag}{AG}, and \textcolor{hag}{HAG}) gradually shift their RD frontiers downward, indicating improved distortion for a given rate. 
\input{figures/simulation/si-recon-len-per-program/fig-recon-len-per-program-trend-beta}
A core prediction of library-based models is that, with experience, learners should construct longer programs by composing previously acquired pieces rather than encoding each subsequence afresh. We index this by subsequence length per subprogram (i.e., ``chunksize') under varying values of the memory-rate parameter~$\beta$ (Fig.~\ref{fig:rd-simulation}(a)).
We fixed all other parameters to~$\lambda_s = 10$, $\lambda_n = 1$, $\aloc = \aglob = 1.0$, $\dglob = \dloc = 0.2$, and imposed no limit on the library sizes. HAG subprograms consistently produced the longest subsequences followed by AG (HAG vs. AG:~$z(99)=1268, p<.001, r=.98, \text{CI}=[.96, .99]$), while Chunking yielded substantially shorter subsequences (HAG vs. Chunking:~$z(99)=1275, p<.001, r=1.0, \text{CI}=[1.0, 1.0]$). 
In Figure~\ref{fig:si:recon-len-per-program-trend-beta}, we plot the same measure separately for different values of $\beta$. As $\beta$~increases, the pressure to minimize description length grows stronger, and HAG (and to a lesser extent AG) responds by calling longer, more compositionally rich library items, yielding progressively longer reconstructed subsequences.  Conversely, at low $\beta$, the rate penalty is weak, distortion dominates the objective, and all models default to conservative, short subprograms that reconstruct local detail accurately. In this regime (e.g., when $\beta=0.1$, HAG does not exceed AG $z(99)=296, p=.99, r=-.53, \text{CI}=[-.79,-.25]$ or PCFG $z(99)=651, p=.45, r=.02, \text{CI}=[-.29,.35]$). 

\input{figures/simulation/si-rd-compute/fig-rd-compute}
\paragraph{Effects of search and backtracking budgets}
Figure~\ref{fig:si:rd-compute}a shows how increasing the expected search budget ($\lambda_s$) influences RD trade-offs across all model classes (darker lines correspond to larger budgets). For all models except from PCFG, larger search budgets systematically improve performance, yielding lower distortion at comparable rates, indicating that broader search enables more effective discovery of high-quality subprograms. 
In PCFG, the apparent advantage when expected search budget $\lambda_s$ is low reflects a \emph{simplicity bias} induced by limited search.
When $\lambda_s$~is small, only few candidates are evaluated during inference, thus ``default'' parses that are low-complexity (i.e., fewer tree expansions) are likely to be accepted. As $\lambda_s$~increases, the model is more likely to propose deeper-nested subprograms or more specific combinations of grammar rules, even though they did not prove to be useful in the past. These unnecessarily complex (relative to the regularities present in the sequence) subprograms can increase rate and also the distortion.
Performance changes between models with or without global libraries. Models with a global library (RLE, AG, HAG) benefit from larger $\lambda_s$, while PCFG, which has no library, is hurt by it. This suggests that an accumulated library acts as a learned proposal distribution---concentrating search on subprograms that have proved useful before, so additional search budget is spent productively rather than on speculative compositions.
Together, these results show that search budget interacts with the structure of the hypothesis space. Without a library, more search means more chances to accept an over-fit composition. Meanwhile, with a library, more search means more chances to recover a genuinely reusable structure. The benefit of computation is therefore conditional on having representations that make broader search worth doing.

Whereas search budget governs how widely the model casts its net for new candidates, backtracking budget governs how readily it revises commitments it has already made. 
Figure~\ref{fig:si:rd-compute}b shows how expected backtracking budget ($\lambda_b$) influences RD trade-offs across all model classes. 
Increasing $\lambda_b$ improves rate–distortion performance for all models, indicating that the ability to revise earlier decisions enables more effective subprogram refinement. 
This effect is especially pronounced for models without local libraries RLE, Chunking, PCFG, AG. In contrast, HAG shows only modest improvements.
 
\input{figures/simulation/si-rd-compute/fig-rd-compute-program}
Figure~\ref{fig:si:computation-program} characterizes how search budget and backtracking budget shape the structural properties of inferred subprograms. Larger search budgets~$\lambda_s$ are associated with increased subprogram depth, indicating how broader exploration helps discover more deeply compositional solutions. 
In contrast, increasing backtracking budget~$\lambda_b$ mainly affects encoding length, as models iteratively revise and elaborate candidate subprograms. Together, these results highlight complementary roles of search and backtracking: search expands the space of compositions considered, while backtracking refines solutions within that space.

\section*{Supplemental Behavioural Analyses}

\subsection*{Melodic sequence statistics}

\input{figures/melody/fig-melody-stats}

As shown in Figure~\ref{fig:si:melody-stats}, the mean segment length per participant was $14.89\pm0.41$~notes. We deliberately chose this range to ensure participants would use some degree of compression, by greatly exceeding canonical working-memory limits \cite{cowan2010magical, oberauer2018benchmarks}. 
The marginal note distribution in the training set was approximately uniform across the six notes, with proportions~$[.170,\,.156,\,.166,\,.167,\,.171,\,.171]$ (maximum deviation from uniform~$<.011$), ensuring that no individual note dominated the material.

\subsection*{Behavioral data}

\input{figures/melody/fig-performance-stats}
Participants’ performance changed systematically across the three experimental phases (Fig.~\ref{fig:si:error-rates-and-reaction-time}). 
Accuracy was quantified as each participant’s average error rate, and response speed as log-transformed reaction time ($\mathrm{LogRT}$). Error rates were lowest during Phase~1 (learning), where participants simply reproduced shown notes. Errors increased substantially in Phase~2 (recall) when they had to reconstruct the melody from memory, and peaked in Phase~3 (predict) when predicting the unseen continuation of the melody. Mean error rates were: $.04 \pm .04$, $.46 \pm .20$, and $.78 \pm .08$ in three phases, respectively. 
Reaction times showed the opposite trend. Participants responded more quickly as the task progressed from learning to recall to prediction: $11.23 \pm 0.37$, $11.04 \pm 0.40$, and $8.74 \pm 0.45$ in the three phases, respectively. 

\input{figures/behavior/si-melody-processed-stats/fig-melody-processed-stats}
Figure~\ref{fig:si:learning-dynamics} characterizes behavioral signatures of learning across repeated exposure to structured melodic sequences. Reaction times and error rates decrease systematically across trials, consistent with increasing fluency and improved memory for repeated structure. Linear mixed-effects analyses reveal progressively faster responses both within and across sequences, indicating learning dynamics over constituent units. 
In parallel, chunk sizes inferred from RT boundaries increase with practice, suggesting that learners gradually integrate smaller elements into larger, more coherent internal representations. 
Differences between \textrm{SegmentID} and \textrm{SequenceID} conditions further indicate that representational organization is shaped by the hierarchical structure of the learning environment.

\subsection*{Self-reports}
Participants’ provided self-reported free-text responses to the post-task survey ``What strategy did you use?''. These reports reflected strong convergence with the model-derived principles of pattern discovery and compositional reuse. Common reported strategies included:
\begin{itemize}
    \item Pattern finding: 
        \textit{“Repeating notes or constant decreases.”} 
        \textit{“I tried to find patterns in the jumps or repetitions.”}
    \item Using melodic structure:
        \textit{“I made up a beat for each note.”}
        \textit{“I imagined a tune in my head.”}
    \item Anchoring on endpoints:
        \textit{“I memorized the first five notes and the last few to see if they repeated.”}
    \item Transforming the stimulus into alternative formats:
        \textit{“I used numbers to remember the sequence.”}
        \textit{“I swayed my body left or right to feel the pattern.”}
    \item Reinforcing visual/auditory cues:
        \textit{“I said the letters out loud to help me remember.”}
        \textit{“I tried to memorize visually.”}
\end{itemize}

Approximately $10\%$ of participants reported that the task felt very difficult, highlighting the cognitive load imposed by sequence recall. Overall, self-reported strategies corroborate the program-learning interpretation: participants spontaneously searched for patterns, reused structural motifs, and formed compact and selective representations of the sequences. These introspective reports align closely with the structured transformations and program-like errors observed in behavioral data.

\section*{Supplemental Plausible Observer Analyses}

\input{figures/behavior/regression/fig-chunking-observer-rt}
In Figure~\ref{fig:behavior-regression}, we showed different explanations in RT jumps from each model. 
AG and PCFG did not explain RT independently of HAG, perhaps because their boundary predictions were moderately correlated with HAG and with each other (AG-HAG: $r=.66$; PCFG-HAG: $r=.58$; AG-PCFG: $r=.67$; Fig.~\ref{fig:si:chunking-observer-rt}a). Accordingly, in the full regression, HAG captured variance most closely associated with hierarchical transition costs, leaving AG negative and PCFG as non-significant predictors. We therefore interpret the AG and PCFG coefficients as reflecting collinearity, rather than evidence that these models predict the opposite behavioral effect.

Chunking showed a qualitatively different relationship to RT. Whereas HAG and RLE predicted increased RT at their model-defined boundaries, Chunking-defined boundaries were associated with faster responses. This pattern is not primarily driven by collinearity, since Chunking was only modestly correlated with the hierarchical predictors (Chunking-HAG: $r=.35$). 
Instead, it reflects how boundaries are defined under Chunking. Because Chunking captures only short, contiguous patterns of reuse, it produces substantially shorter subsequences than the hierarchical models (Fig.~\ref{fig:si:chunking-observer-rt}b) and therefore places boundaries far more densely. 
The majority of these extra boundaries fall in the interior of what HAG treats as a single subsequence.
Thus, two structural factors explain the resulting negative coefficient. 
First, subsequence length itself is positively associated with $\mathrm{LogRT}$ in a mixed-effects regression $(\gamma=0.028, z=3.01, p=.003, \mathrm{CI}=[.010, .047]$; Fig.~\ref{fig:si:chunking-observer-rt}c), such that transitions between larger units are slower regardless of which model defines them. 
Second, the regressor for the Chunking model is dominated by positions that are within a single HAG subsequence, with RTs that are relatively fast. This results in a generally negative relationship between Chunking-defined boundaries and RTs, which tend to be faster in the aggregate.

\section*{Supplemental Modeling Analyses}

\input{figures/model/si-fitting/fig-model-fitting-score}
Across participants, HAG provided consistently better fits than alternative accounts based on it achieves higher log-likelihoods (HAG:~$-1309.78 \pm 48.63$, AG:~$-1668.30 \pm 63.27$, PCFG:~$-2607.41 \pm 119.41$, Chunking:~$-2595.45 \pm 105.71$, RLE:~$-2576.90 \pm 110.61$) while also yielding lower BIC values (HAG:~$2649.52 \pm 97.26$, AG:~$3360.56 \pm 126.54$, PCFG:~$5232.80 \pm 238.83$, Chunking:~$5214.87 \pm 211.43$, RLE:~$5177.77 \pm 221.23$).
HAG was the most frequently top-ranked model at the participant level, winning for $53/96$ participants, followed by AG ($21/96)$ and PCFG ($17/96$); RLE and Chunking each won for a small minority ($3/96$ and $2/96$, respectively). Thus, while a meaningful fraction of participants were best described by simpler accounts - most often AG or PCFG - HAG remained the modal best-fitting model and dominated overall in both likelihood and BIC.


\input{figures/model/si-recovery/fig-model-recovery}
To assess identifiability, we conducted a model-recovery analysis in which synthetic datasets were generated from each candidate model and then refit with the full model set, using the same likelihood and fitting pipeline as in the main analyses. We first fit each model to participants’ data and then defined an empirical parameter range for recovery using the distribution of best-fitting parameter values observed across participants. For each generating model, we sampled parameters from this empirically grounded range, simulated datasets matched to the experimental design, and applied the same model-comparison criterion to assign a recovered label. This approach targets practical identifiability in the regime actually supported by the data (i.e., where fitted parameters tend to lie, rather than identifiability under arbitrary parameter settings that participants never exhibit).

Model-recovery analyses indicate that each candidate model is generally identifiable. From the confusion matrix, $P(\text{recovered} | \text{true})$, recovery is strongest for HAG ($.98$) and AG ($.88$), with moderate diagonal recovery for PCFG ($.73$), Chunking ($.72$), and RLE ($.68$). Misclassifications primarily reflect a bias towards AG overfitting data generated by PCFG, Chunking and RLE. In contrast,  between HAG and other models is rare (only $.02$ of HAG datasets recovered as AG).
The complementary inversion matrix, $P(\text{true} | \text{recovered})$, further shows that when PCFG, Chunking, or RLE is the winning model, the label is highly diagnostic of the true generating model. Recovered AG and HAG labels are somewhat less exclusive, reflecting overlap in the patterns they can produce, though HAG remains substantially more diagnostic than AG ($. 76$ vs. $.55$).

\input{figures/model/si-recovery/fig-param-recovery} 
Beyond model identifiability, we also assessed whether HAG's individual parameters can be reliably recovered. We treated each participant’s fitted HAG as a data-generating model, simulated synthetic choices using their estimated parameters, and then applied the same fitting pipeline to recover parameters from the simulated data. As shown in Fig.~\ref{fig:si:param-recovery}, recovered values closely track the generating values and preserve participants’ rank ordering (Kendall’s~$\tau$ shown in each panel; for memory resource $\beta^{-1}$, $\tau=.73, p<.001$; for expected search budget $\lambda_s$, $\tau=.61, p<.001$; for expected backtracking budget $\lambda_b$, $\tau=.67, p<.001$; for global reuse $\aglob$, $\tau=.56, p<.001$; for local reuse $\aloc$, $\tau=.58, p<.001$), supporting the identifiability of the parameters used in subsequent analyses. 

Finally, we examined whether the fitted parameters might be redundant with one another. For instance, whether high values of one parameter could be systematically compensated by low values of another, which would undermine the interpretability of individual parameters even if each can be recovered in isolation. Across participants, pairwise correlations among fitted parameters were weak (largest magnitudes $\sim .20-.25$; Fig.~\ref{fig:si:model-recovery}c), indicating that the five parameters capture distinct sources of variation rather than collinear axes of the same underlying construct. This complements the parameter-recovery analysis above. Recovery shows that each parameter can be estimated reliably, while the low cross-parameter correlations show that the parameters are not silently doing each other's work.

\input{figures/model/si-fitting/fig-ph3pred}
In Phase~3, participants generated free continuations of the melody they observed in segments 1 to 5, allowing us to test whether the libraries inferred from earlier phases generalize out-of-sample. For each participant, we fit HAG to their behavior up to the onset of Phase~3 , yielding an individualized library that can be used as a generative model to sample candidate continuations. 
The main analysis (Fig.~\ref{fig:model-fitting}c-e) established that HAG provides the best per-trial likelihood of participants' continuations and the strongest participant-level correlation between observed and predicted error rates. Here we report a complementary distributional check: rather than asking how closely HAG's predictions track each participant's exact continuation, we ask whether HAG-generated continuations look like participant continuations as samples from the same population. This is a stricter test of the generative claim---a model could in principle achieve high likelihood by hedging probability mass across many plausible continuations without ever producing trajectories that themselves resemble human continuations. As shown in Figure~\ref{fig:si:ph3pred}, the error-rate distributions of HAG-generated and participant-produced continuations closely overlapped, and the per-participant discrepancy between $\mathrm{ErrorRate}_{\mathrm{HAG}}$ and $\mathrm{ErrorRate}_{\mathrm{Pressed}}$ was concentrated near zero. Participant-specific HAG libraries thus pass both tests: they assign high likelihood to the specific continuations participants produced and generate continuations whose distributional properties match those participants at the population level.

\input{figures/model/si-fitting/fig-simulation-learner}

Beyond relating fitted parameters to aggregate behavioral measures, we asked whether the fitted models captured graded variation in behavior across segments. For each participant, we simulated the fitted model on the same set of sequences encountered in the experiment and, for each segment, derived a predicted error-rate and a predicted chunk-size measure. We compared these model-derived quantities with the corresponding observed measures: actual error rate and chunk size inferred from RT.
By fitting mixed-effects regressions in which the empirical measure was predicted jointly by the simulated predictions from all candidate models, the regression line shown for HAG in Figure~\ref{fig:si:simulation-learner} represents its partial effect after controlling for competing model predictions. To complement this analysis, Figure~\ref{fig:si:simulation-learner} also reports participant-level correlations between actual and simulated values for each model (bar plots on the right of each panel).

HAG explained significant unique variance in observed error rate, indicating that the fitted model captured graded differences in behavioral difficulty across segments (HAG: $r=.84, p<.001$; AG: $r=.75, p<.001$; PCFG: $r=.70, p<.001$; Chunking: $r=.71, p<.001$; RLE: $r=.50, p<.001$). 
The same pattern held for chunk size inferred from RT, indicating that HAG also tracked variation in the granularity of participants' internal representations (HAG: $r=.30, p<.001$; AG: $r=.15, p<.13$; PCFG: $r=.07, p<.46$; Chunking: $r=.26, p<.05$; RLE: $r=.13, p<.18$). 

\input{figures/model/si-hag-param-dist/fig-param-dist}
Figure~\ref{fig:si:fitted-param-distribution} plots the marginal distributions of participant-level maximum-likelihood parameter estimates for HAG. The figure serves three purposes. First, it confirms that fitted estimates populate the interior of the search grid for all five parameters, with relatively few participants pinned at boundary values, indicating that the grid was wide enough to accommodate the empirical range. Second, the distributions reveal substantial between-participant heterogeneity along every dimension---memory budget $\left(\beta^{-1}\right)$, expected search and backtracking budgets ( $\lambda_s, \lambda_b$ ), and the strength of library reuse at global versus local scales $\left(\alpha_{\text {glob }}, \alpha_{\text {loc }}\right)$---establishing that participants do not cluster around a single canonical strategy and that the model has identified meaningful variation rather than collapsing onto a population-level prototype. Third, this empirical heterogeneity is the substrate for the individual-differences analyses reported in the main text: the relationships between fitted parameters and behavioral measures (error rate, chunk size, reuse-parameter contrast) are well-defined precisely because participants occupy distinct regions of the parameter space.

%% file: figures/simulation/rd-simulation/algorithm.tex
\begin{algorithm}[!th]
\caption{Approximate inference with hierarchical $\mathrm{HPYP}$ program libraries}
\label{alg:inference}
\small
\KwIn{Sequences $\{X^{(i)}\}_{i=1}^N$, base grammar $H$, $\mathrm{HPYP}$ hyperparameters, $\beta$, $\lambda_s$, $\lambda_b$, $\tau$.}
\KwOut{Inferred subprograms $\{\pi^{(i)}_j\}$ and updated libraries.}

Initialize global library $G_0$\;

\For{$i \leftarrow 1$ \KwTo $N$}{
  Initialize local library $G_i$ and stack $\mathcal{S}\leftarrow[\,]$\;
  $t \leftarrow 1$\;
  
  \While{$t \le |X^{(i)}|$}{
    $\mathcal{C} \leftarrow \emptyset$\;
    
    \tcp{Propose up to $N_s \sim \operatorname{Pois}\left(\lambda_s\right)$ candidate subprograms}
    \For{$r \leftarrow 1$ \KwTo $N_s$}{
      Sample source $z \sim \texttt{HPYPSource}(G_i,G_0,H)$\;
      Sample candidate $\pi$ from source $z$\;
      \If{$\pi$ is well-typed}{
        $\hat{X}_{\pi} \leftarrow g(\pi)$\;
        $m_{\pi} \leftarrow |\hat{X}_{\pi}|$\;
        \If{$t+m_{\pi}-1 \le |X^{(i)}|$}{
          $\mathcal{C} \leftarrow \mathcal{C} \cup \{\pi\}$\;
        }
      }
    }

    \tcp{Score candidates on the span they induce}
    \ForEach{$\pi \in \mathcal{C}$}{
      $\hat{X}_{\pi} \leftarrow g(\pi)$\;
      $m_{\pi} \leftarrow |\hat{X}_{\pi}|$\;
      $X^{(i)}_{\pi} \leftarrow X^{(i)}_{t:t+m_{\pi}-1}$\;
      $\mathcal{L}(\pi) \leftarrow d\!\left(X^{(i)}_{\pi},\hat{X}_{\pi}\right)+\beta\,L(\hat{X}_{\pi})$\;
    }

    \tcp{Select and commit}
    Sample $\pi^\star \in \mathcal{C}$ with probability
    $p(\pi^\star)\propto \exp[-\mathcal{L}(\pi^\star)/\tau]$\;
    Push $\pi^\star$ onto $\mathcal{S}$\;
    
    \tcp{Optional bounded backtracking}
    \For{$b \leftarrow 1$ \KwTo $\min(N_b \sim \operatorname{Pois}\left(\lambda_b\right),|\mathcal{S}|)$}{
      Resample the last $b$ subprograms and accept the revision if it improves the total posterior score\;
    }

    Update $G_i$ with the accepted subprogram(s)\;
    $t \leftarrow t + |g(\pi^\star)|$\;
  }

  Update $G_0$ with accepted subprograms from sequence $i$\;
}

\Return{$\{\pi^{(i)}_j\}$ and updated libraries}\;
\end{algorithm}

%% file: figures/simulation/si-eval/fig-eval-rd-curve.tex
\begin{figure}[!t]
    \centering
    \includegraphics[width=0.7\linewidth]{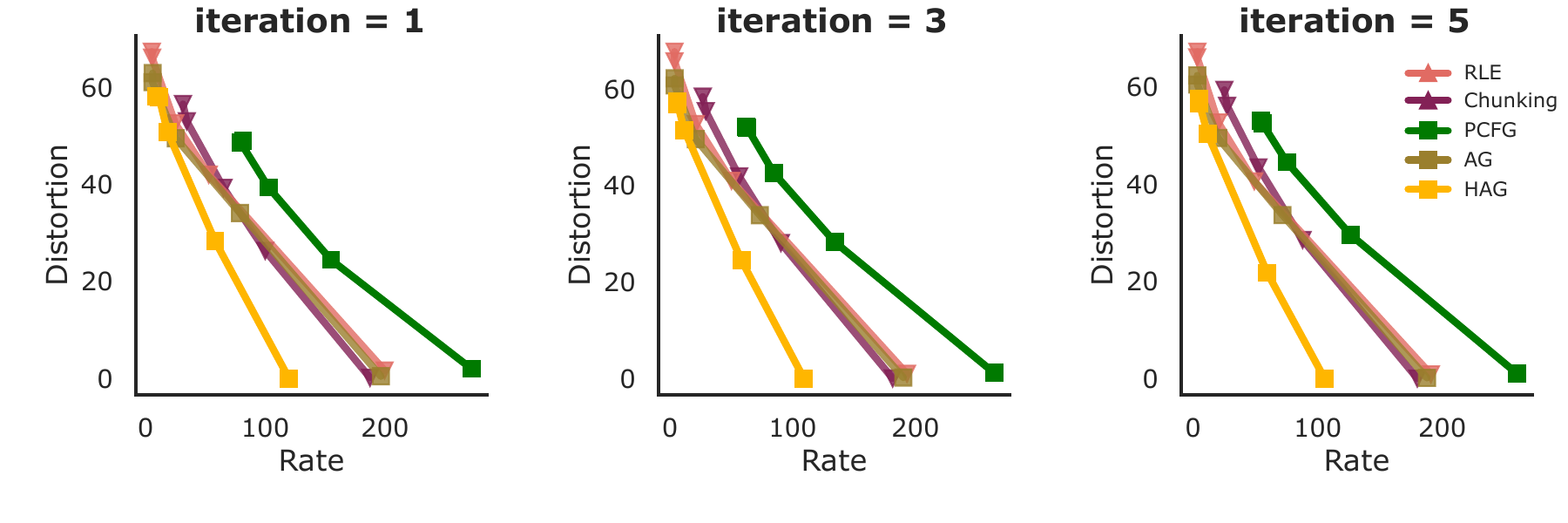}
    \caption{\textbf{Generalization rate–distortion (RD) performance.} 
    One-shot, three-shot, and five-shot generalization RD curves obtained by compressing unseen sequences using the trained library without additional optimization iterations. 
    All results use~$\lambda_s = 10$, $\lambda_b = 1$, $\alpha_\mathrm{loc} = \alpha_\mathrm{glob} = 1.0$, and~$d_\mathrm{loc} = d_\mathrm{glob} = 0.2$. The library was trained on $50$~sequences, and the curves report compression performance on held-out sequences.}
    \label{fig:si:eval-rd-curve}
\end{figure}

%% file: figures/simulation/si-rd-over-time/fig-rd-over-time.tex
\begin{figure}[!t]
    \centering
    \includegraphics[width=1.0\linewidth]{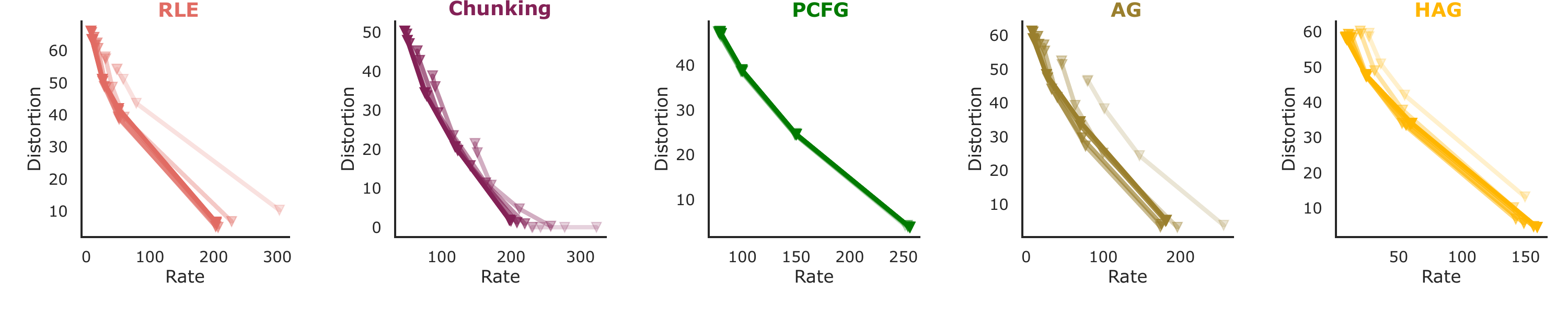}
    \caption{\textbf{Rate–distortion (RD) trajectories during training across models.}
    RD curves over the course of training for each model.
    Within each panel, darker lines correspond to more training data ($2\%$, $10\%$, $40\%$, $80\%$, and $100\%$, for $\lambda_s=10$ and $\lambda_b=1$, respectively). Across models (except PCFG which lacks a library), increased training data shifts RD curves toward lower distortion at equivalent rates.
    }
    \label{fig:si:rd-over-time}
\end{figure}

%% file: figures/simulation/si-recon-len-per-program/fig-recon-len-per-program-trend-beta.tex
\begin{figure}[!t]
    \centering
    \includegraphics[width=1.0\linewidth]{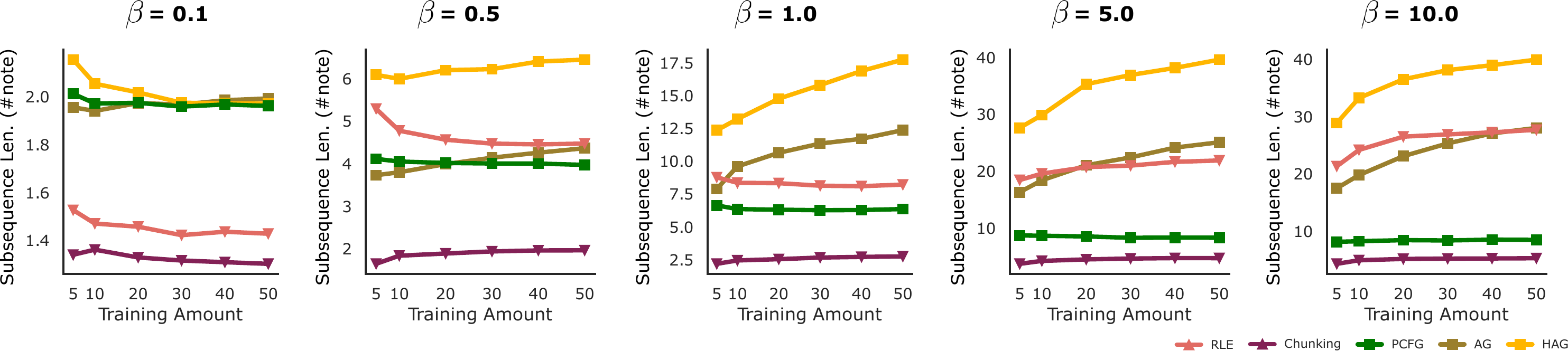}
    \caption{
    \textbf{Reconstructed subsequence length under varying memory constraints ($\beta$)}.
    Average subsequence length per selected subprogram (in notes) as a function of training data size, shown separately for different values of the memory--rate parameter $\beta$. 
    As $\beta$ increases, AG and especially HAG produce progressively longer reconstructed subsequences. 
    With additional training data, this trend becomes stronger for AG and HAG, whereas Chunking, PCFG, and RLE change comparatively little.}
    \label{fig:si:recon-len-per-program-trend-beta}
\end{figure}

%% file: figures/simulation/si-rd-compute/fig-rd-compute.tex
\begin{figure}[!t]
    \centering
    \includegraphics[width=1.0\linewidth]{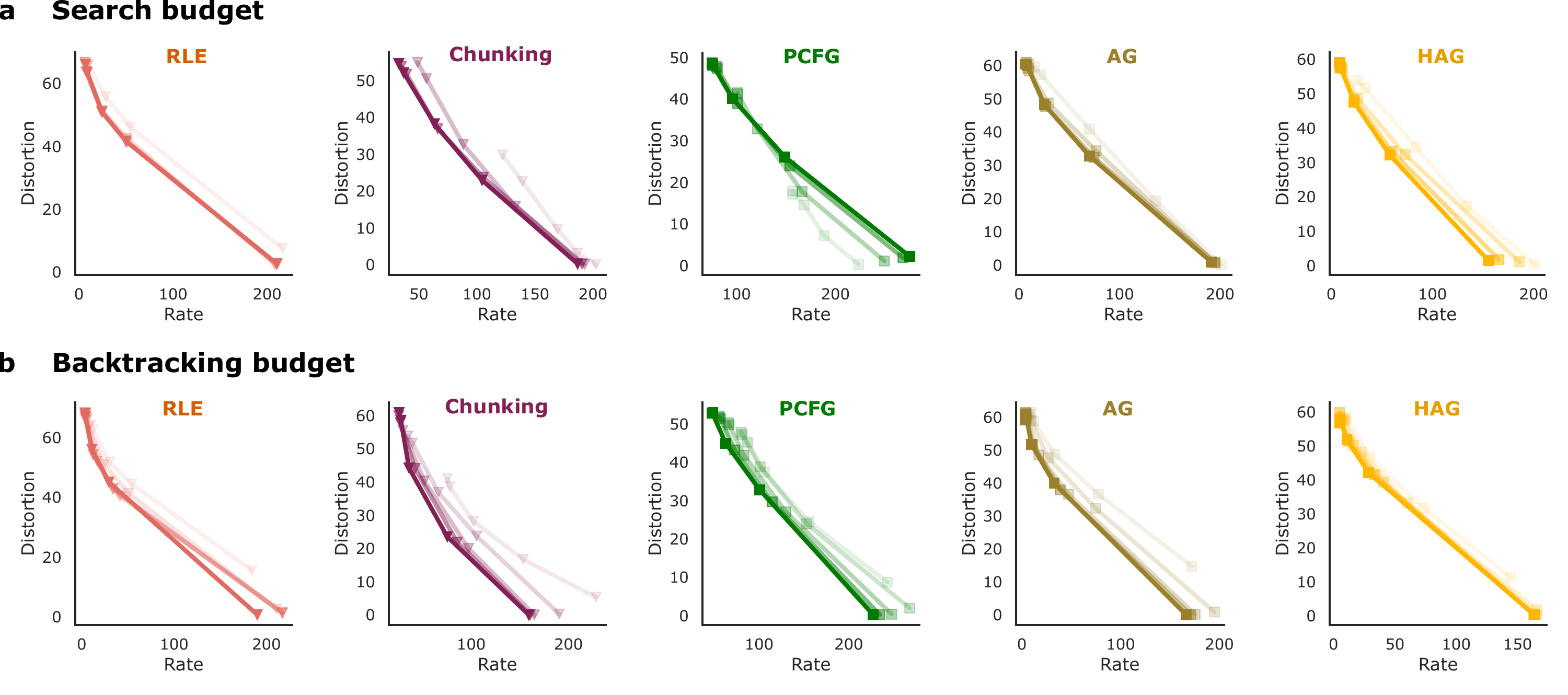}
    \caption{\textbf{Rate–distortion (RD) curves across models under varying computational resources.} \textbf{(a)} expected search budgets $\lambda_s$ and \textbf{(b)} expected backtracking budgets~$\lambda_b$. 
    Within each panel, darker lines correspond to larger budgets ($\lambda_s \in [1,5,10,20]$; $\lambda_b \in [0,1,3,5,10]$).}
    \label{fig:si:rd-compute}
\end{figure}

%% file: figures/simulation/si-rd-compute/fig-rd-compute-program.tex
\begin{figure}[!t]
    \centering
    \includegraphics[width=0.52\linewidth]{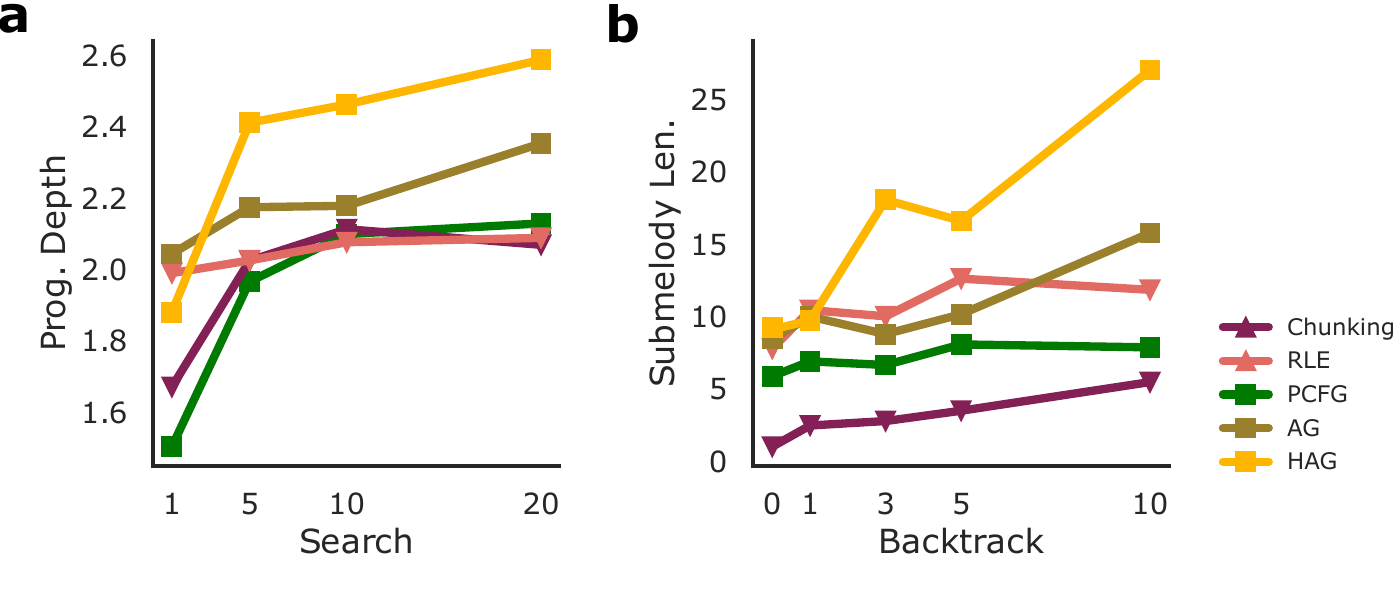}
    \caption{\textbf{Effects of search budget and backtracking budget on subprogram structure.}
    \textbf{(a)} Average subprogram depth as a function of expected search budget ($\lambda_s=1,5,10,20$). 
    \textbf{(b)} Average subsequence length as a function of expected backtracking budget ($\lambda_b=0,1,3,5,10$). 
    Increasing search budget enables the discovery of deeper compositional subprograms, while increasing backtracking budget leads to longer encodings, reflecting iterative refinement over multiple revisions.}
    \label{fig:si:computation-program}
\end{figure}

%% file: figures/melody/fig-melody-stats.tex
\begin{figure}[!t]
    \centering
    \includegraphics[width=0.68\textwidth]{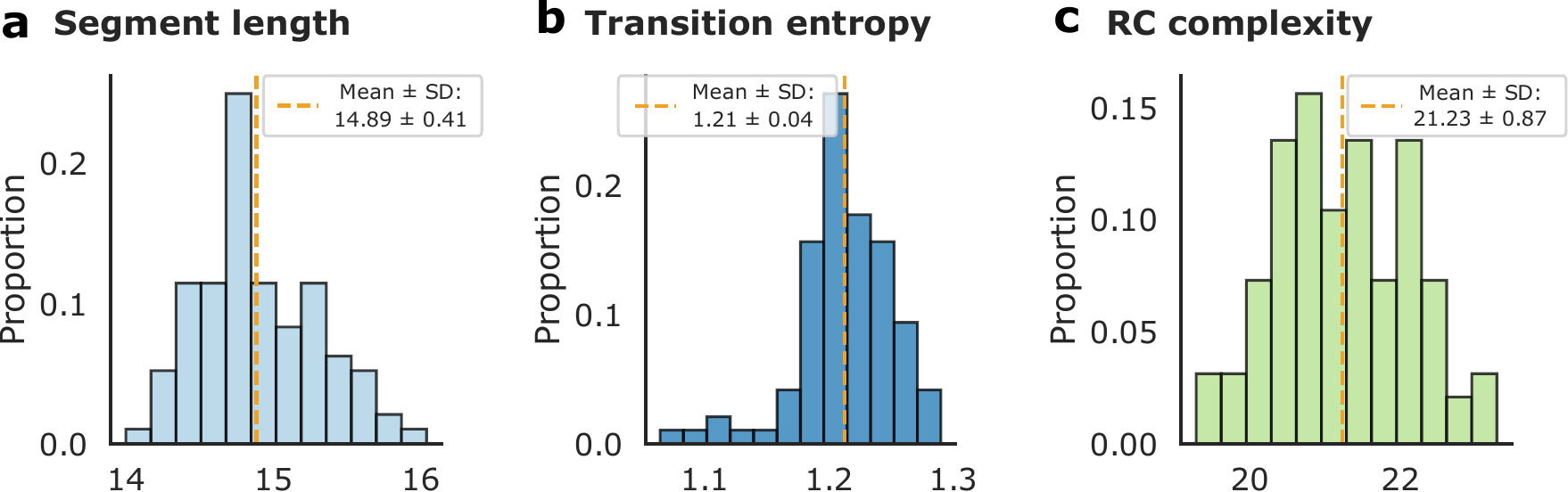}
    \caption{\textbf{Distributions of melody statistics across participants.}
    Each panel shows the histogram across participants of the participant-wise mean for the indicated measure. Dashed vertical lines mark the across-participant mean; shaded bands indicate $\pm$1~s.d. Summary values: (\textbf{a}) $M=14.89\pm0.41$~notes; (\textbf{b}) $H_1=1.21 \pm 0.04$~bits/note; (\textbf{c}) $RC=21.23\pm0.87$~symbols~\cite{dehaene2022symbols}.}
    \label{fig:si:melody-stats}
\end{figure}

%% file: figures/melody/fig-performance-stats.tex
\begin{figure}[!t]
    \centering
    \includegraphics[width=0.68\textwidth]{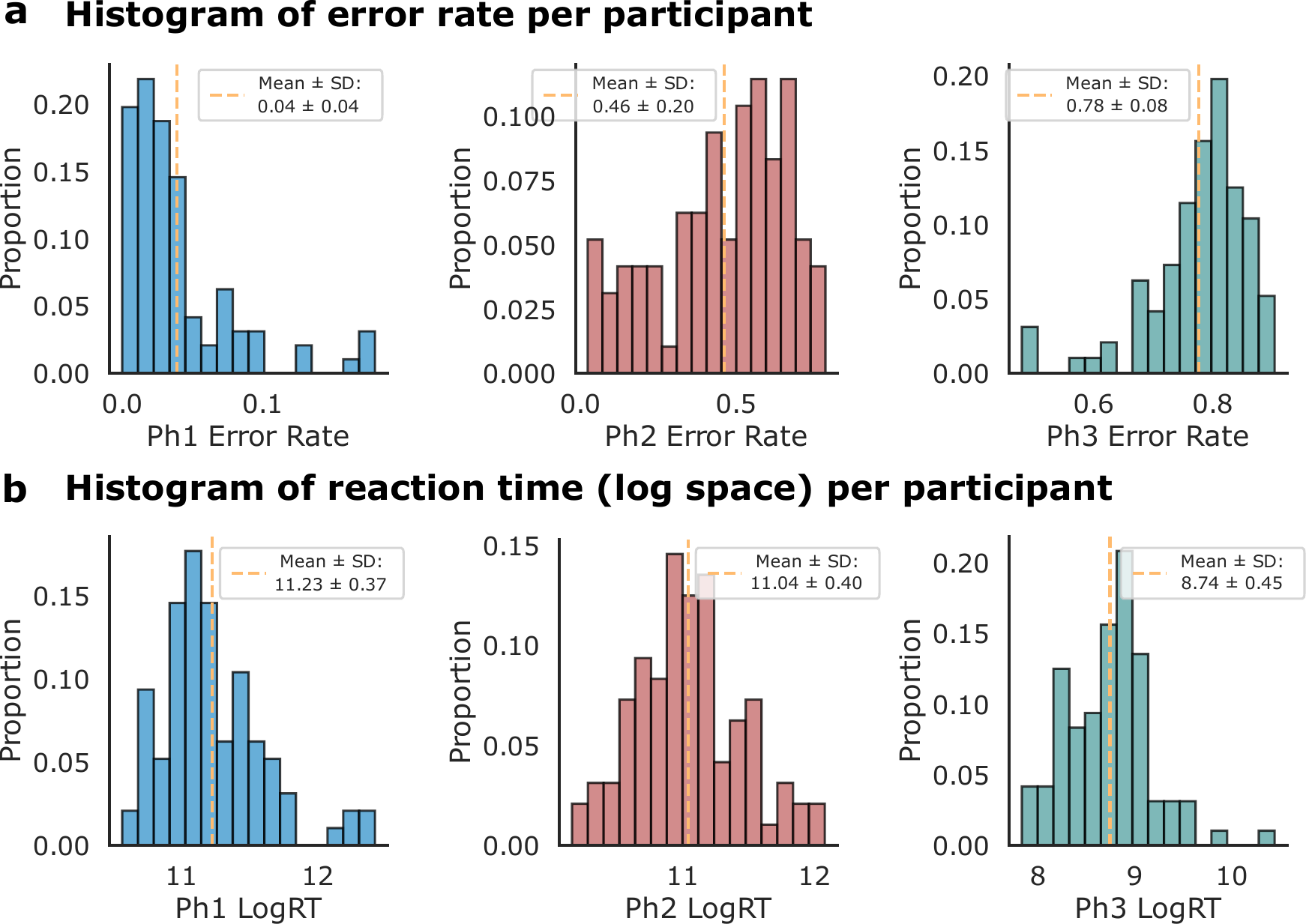}
    \caption{\textbf{Overall behavioral performance across the three experimental phases. }
    \textbf{(a)} Distributions of per-participant error rates in each phase. Vertical dashed lines indicate phase-wise means.
    \textbf{(b)} Distributions of per-participant reaction times (log-transformed). Participants respond progressively faster across phases, with the fastest responses during prediction. Vertical dashed lines indicate phase-wise means. 
    }
    \label{fig:si:error-rates-and-reaction-time}
\end{figure}

%% file: figures/behavior/si-melody-processed-stats/fig-melody-processed-stats.tex
\begin{figure}[!t]
    \centering
    \includegraphics[width=0.72\linewidth]{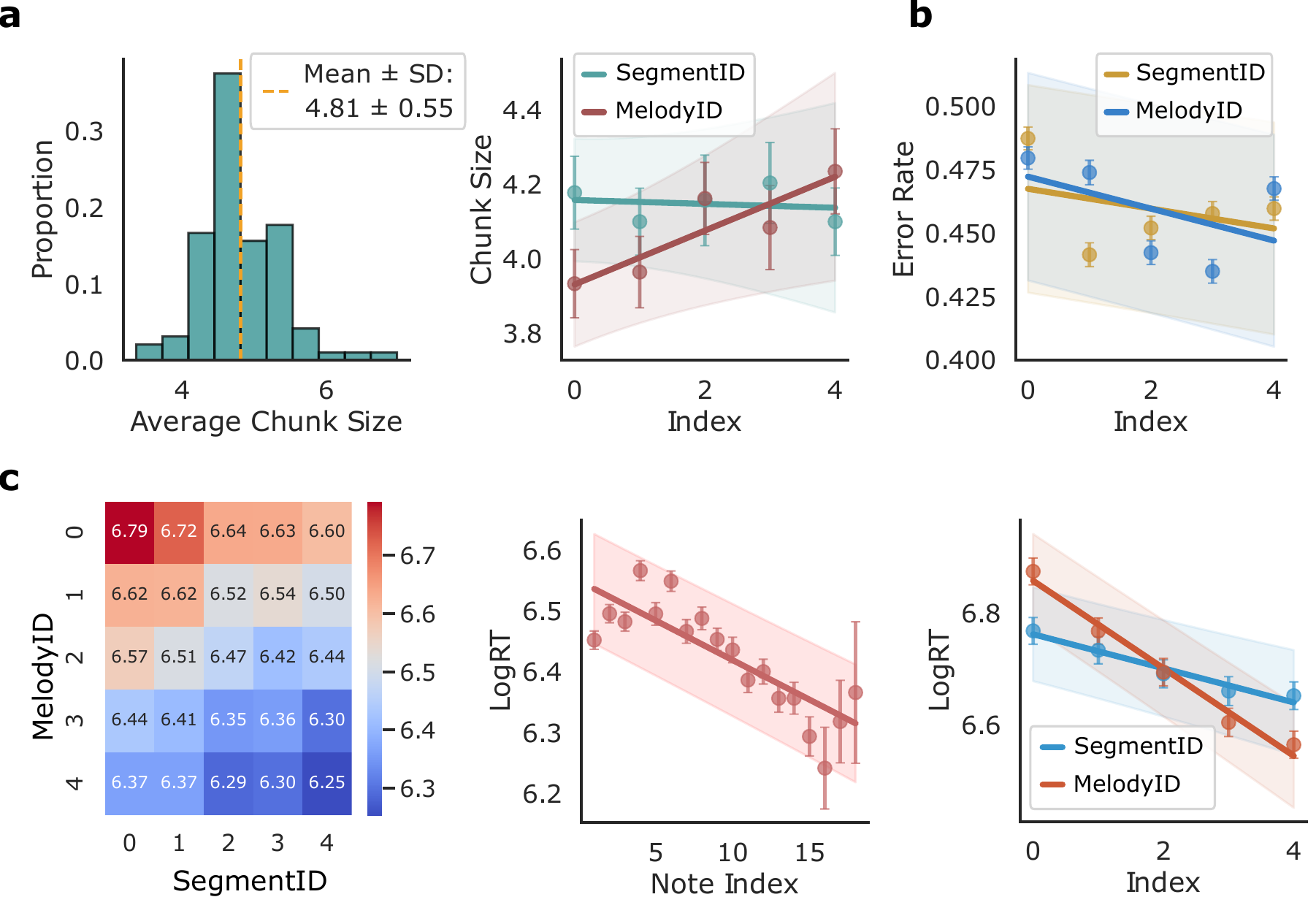}
    \caption{
    \textbf{Behavioral learning dynamics across segments and sequences.}
    \textbf{(a)} Left: Distribution of average chunk sizes inferred from participants’ inter-key timing during Phase 2, with dashed line indicating mean $\pm$ s.d. 
    Right: Estimated chunk size as a function of segment index for \textrm{SegmentID} and \textrm{SequenceID} conditions (points denote means; error bars indicate s.e.m.).
    \textbf{(b)} Error rate as a function of exposure index for \textrm{SegmentID} and \textrm{SequenceID} conditions, with linear trend estimates shown.
    \textbf{(c)} Left: Heat map of mean log-transformed reaction times ($\mathrm{LogRT}$) across segments and sequences. 
    Middle: Linear mixed-effects regression of $\mathrm{LogRT}$ against note index within segments during Phase 2. 
    Right: $\mathrm{LogRT}$ as a function of exposure index for \textrm{SegmentID} and \textrm{SequenceID} conditions.
    Across measures, reaction times and error rates decrease with experience, while inferred chunk sizes increase, indicating progressive learning, improved fluency, and the formation of larger internal units over repeated exposure.}
    \label{fig:si:learning-dynamics}
\end{figure}

%% file: figures/behavior/regression/fig-chunking-observer-rt.tex
\begin{figure*}[!t]
    \centering
    \includegraphics[width=0.92\linewidth]{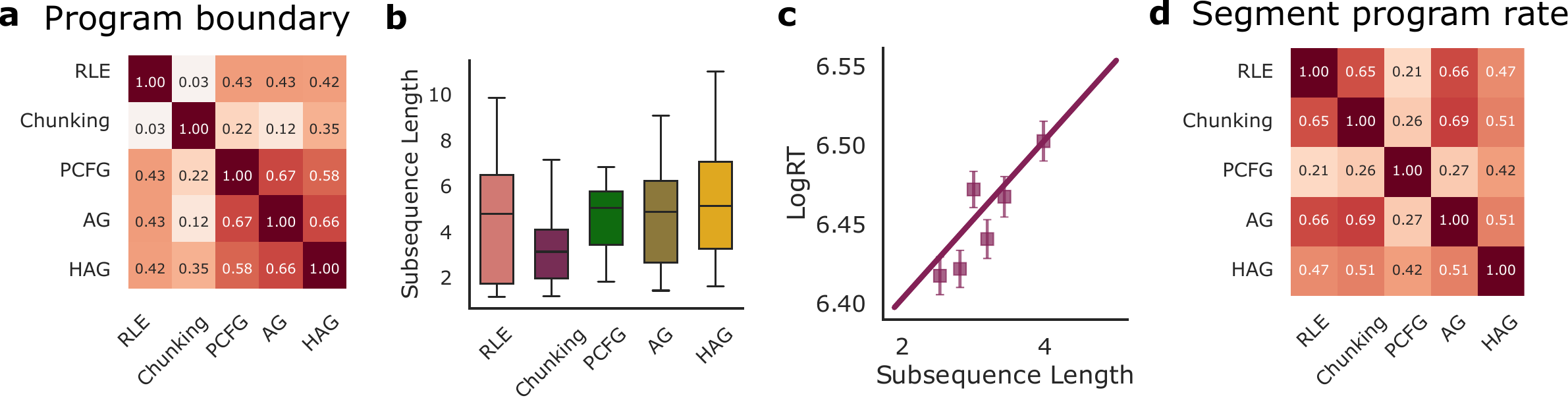}
    \caption{
    \textbf{Correlation structure and representational scale of plausible-observer-based predictors.}
    \textbf{(a)} Mean within-participant correlations among model-predicted boundary signals.
    \textbf{(b)} Distribution of subsequence lengths inferred by each model. Chunking produced shorter subsequences overall than other models.
    \textbf{(c)} Relationship between subsequence length and $\log \mathrm{RT}$ for the Chunking model. Longer Chunking-defined subsequences were associated with slower responses, consistent with a mixed-effects regression showing that subsequence length positively predicted $\log \mathrm{RT}$($\gamma=.028$, $z=3.01$, $p=.003$, $\mathrm{CI}= [.010,.047])$. 
    \textbf{(d)} Mean within-participant correlations among model-based segment-complexity. 
    }
    \label{fig:si:chunking-observer-rt}
\end{figure*}

%% file: figures/model/si-fitting/fig-model-fitting-score.tex
\begin{figure}[!t]
    \centering
    \includegraphics[width=0.75\textwidth]{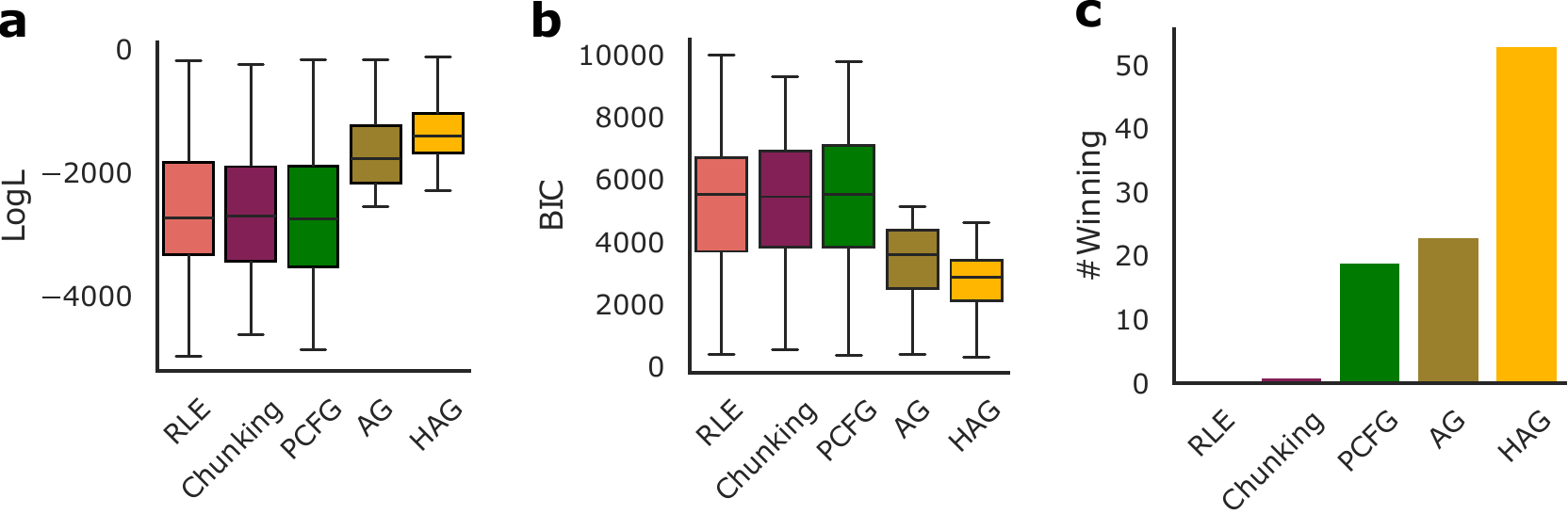}
    \caption{
    \textbf{Model-fit comparison across candidate models. }
    Boxplots summarize the distribution across participants of \textbf{(a)} maximized log-likelihood (LogL; higher indicates better fit).
    \textbf{(b)} Bayesian Information Criterion (BIC; lower indicates better fit after penalizing model complexity).
    \textbf{(c)} Bars show the number of participants (among $96$ participants in total) for which each model achieved the lowest BIC (``wins''). 
    Boxes indicate the median and interquartile range; whiskers extend to $1.5 \times IQR$; points denote outliers.
    }
    \label{fig:si:model-fitting-score}
\end{figure}

%% file: figures/model/si-recovery/fig-model-recovery.tex
\begin{figure}[!t]
    \centering
    \includegraphics[width=0.8\textwidth]{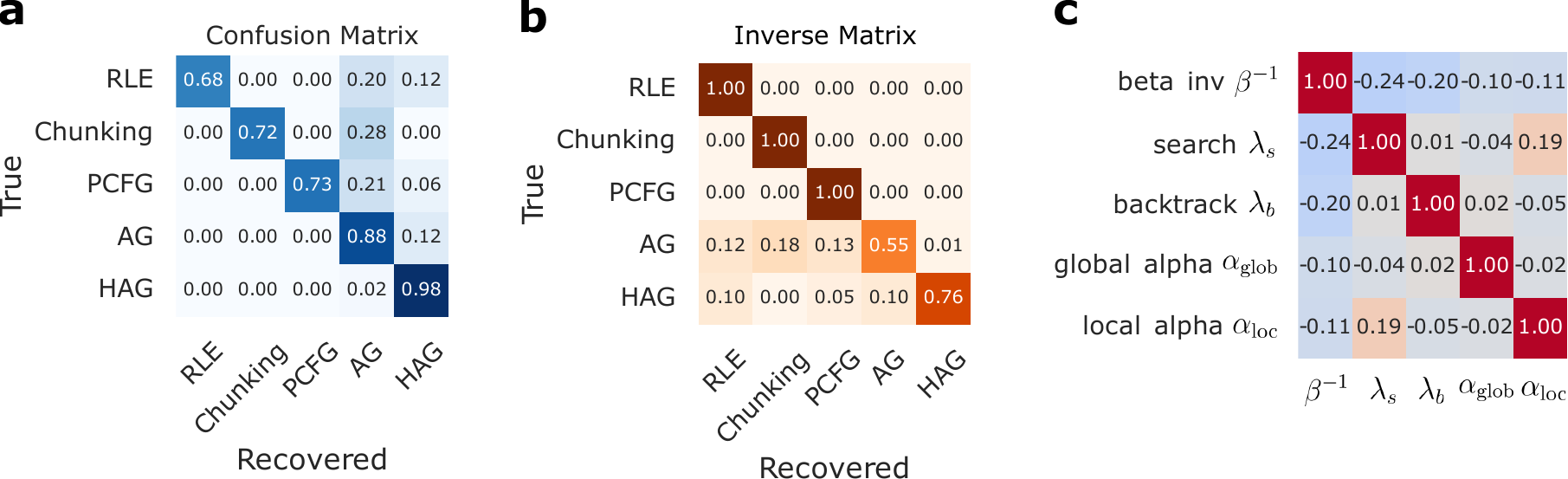}
    \caption{
    \textbf{Model recovery and parameter interdependence. }
    \textbf{(a)} Model-recovery confusion matrix. Rows indicate the true data-generating model and columns indicate the recovered model. Cell values report the proportion of synthetic datasets assigned to each recovered model (row-normalized).
    \textbf{(b)} Inverse matrix. Rows indicate the true model and cell values are normalized by recovered model (column-normalized), showing $P$(
    true model $\mid$ recovered model) to quantify how diagnostic each recovered label is of the generating process.
    \textbf{(c)} Across-participant parameter correlations. Heat map shows pairwise Pearson correlations between fitted parameter estimates across participants (inverse beta, search, backtracking, global alpha, local alpha); diagonal entries are 1 by definition. 
    }
    \label{fig:si:model-recovery}
\end{figure}

%% file: figures/model/si-recovery/fig-param-recovery.tex
\begin{figure}[!t]
    \centering
    \includegraphics[width=\linewidth]{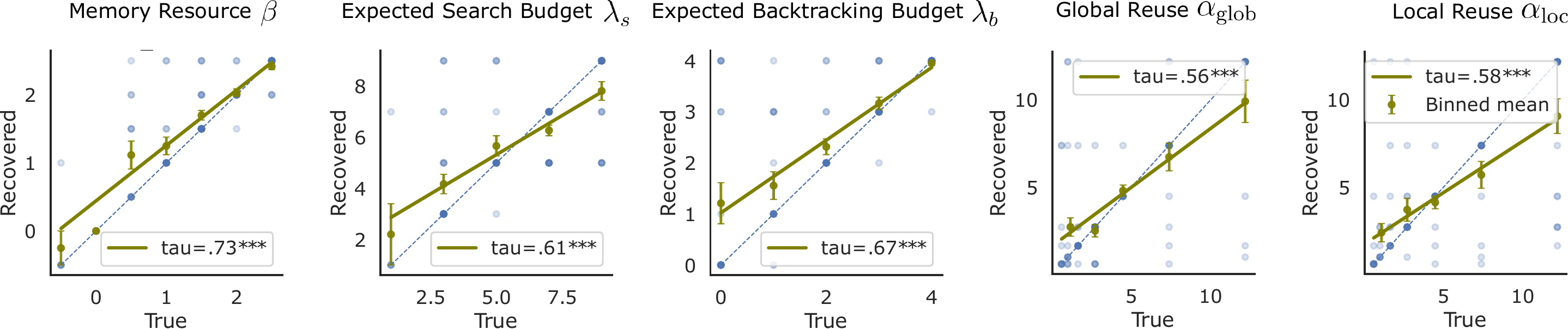}
    \caption{\textbf{Parameter recovery for HAG. }
    Each panel plots recovered estimates against the ground-truth generating values for one parameter: inverse memory resource constraint~$\beta^{-1}$, expected search budget~$\lambda_s$, expected backtracking budget~$\lambda_b$, global reuse~$\aglob$, and local reuse~$\aloc$. Points correspond to participants; dashed line indicates perfect recovery (identity line). Reported Kendall's $\tau$ summarizes rank-order correspondence between true and recovered parameters (all ${ }^{* * *} p<.001$ ).
    }
    \label{fig:si:param-recovery}
\end{figure}

%% file: figures/model/si-fitting/fig-ph3pred.tex
\begin{figure}[!t]
    \centering
    \includegraphics[width=0.46\linewidth]{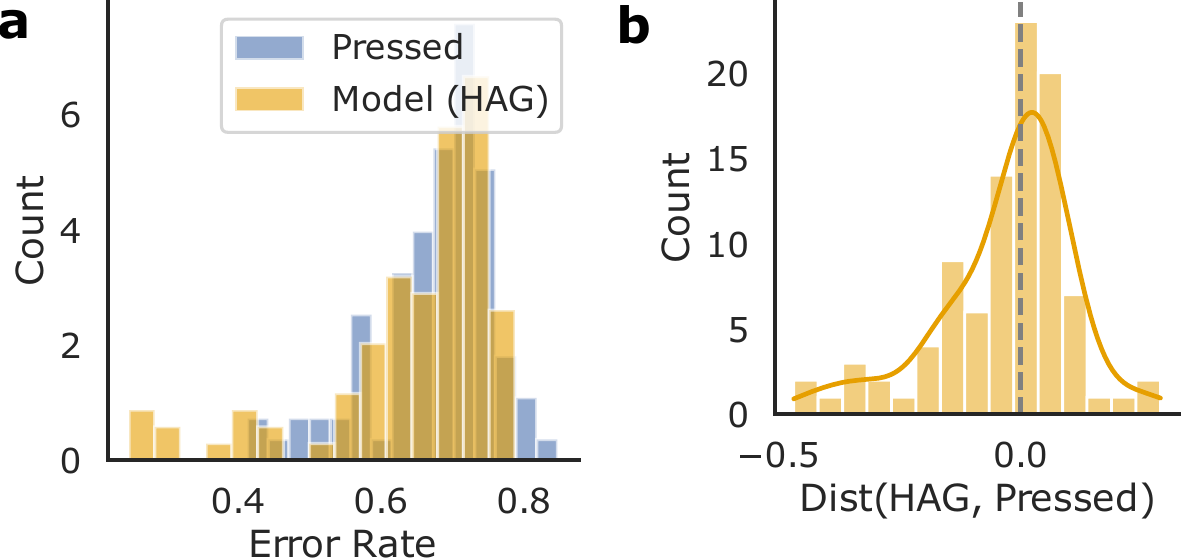}
    \caption{\textbf{Distributional match between human Phase-3 composition errors and HAG-generated continuations. }
    \textbf{(a)} Histogram of Phase-3 error rates (note-wise mismatch to the held-out ground-truth continuation) for participants' pressed continuations (blue) and continuations sampled from each participant's fitted HAG libraries (orange). 
    \textbf{(b)} Histogram of the per-participant discrepancy between model and human error rates, Dist(HAG, Pressed) $=$ ErrorRate $_{\text {HAG }}-$ ErrorRate $_{\text {Pressed }}$; dashed line marks zero (perfect match). Overall, the discrepancy distribution is centered near zero, indicating that participant-specific HAG libraries reproduce the population-level variability in composition accuracy.
    }
    \label{fig:si:ph3pred}
\end{figure}

%% file: figures/model/si-fitting/fig-simulation-learner.tex
\begin{figure}[!t]
    \centering
    \includegraphics[width=0.98\linewidth]{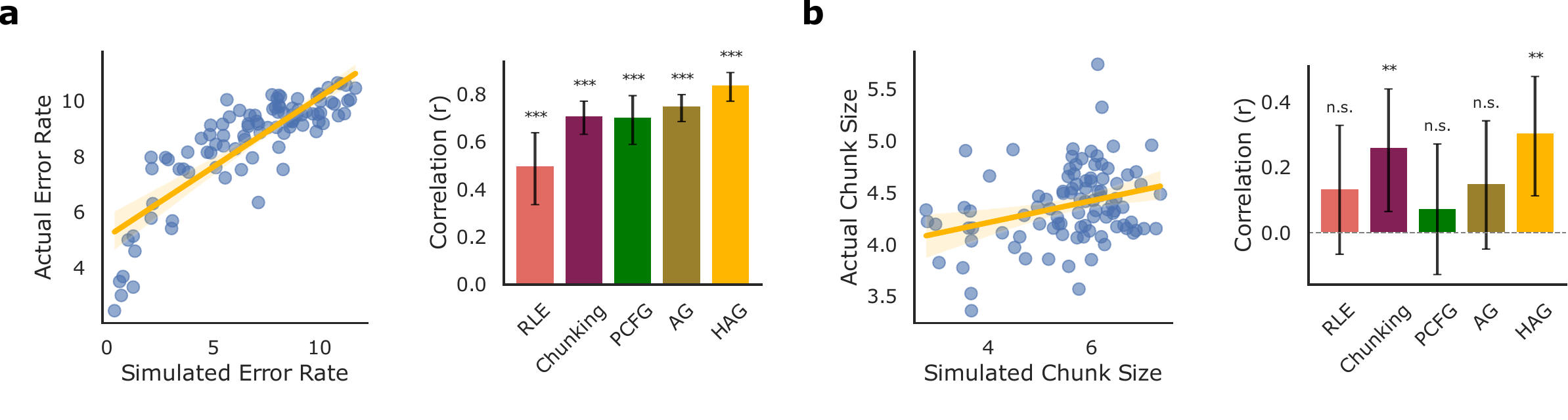}
    \caption{\textbf{Simulation-based prediction of error rate and chunk size.}
    \textbf{(a)} Error rate. Left, partial effect of the HAG predictor from a mixed-effects regression relating actual error rate to simulated error rates from all candidate models jointly, with participant included as a random effect ($\gamma=.760$, $z=4.47$, $p<.001$, $\text{CI}=[.427, 1.092]$;). 
    Right, mean Pearson correlation between simulated and actual error rates across participants for each model.
    \textbf{(b)}, Chunk size. Left, partial effect of the HAG predictor from a mixed-effects regression relating actual chunk size, inferred from RT, to simulated chunk size from all candidate models jointly ($\gamma=.151$, $z=2.101$, $p<.05$, $\text{CI}=[.010, .292]$). Simulated chunk size was derived from the length of subsequences. Right, mean Pearson correlation between simulated and actual chunk sizes across participants for each model. 
    Each scatter point corresponds to one participant. The line shows the fitted relationship for HAG after controlling for competing models. Error bars denote 95\% confidence intervals. 
    }
    \label{fig:si:simulation-learner}
\end{figure}

%% file: figures/model/si-hag-param-dist/fig-param-dist.tex
\begin{figure}[!t]
    \centering
    \includegraphics[width=\linewidth]{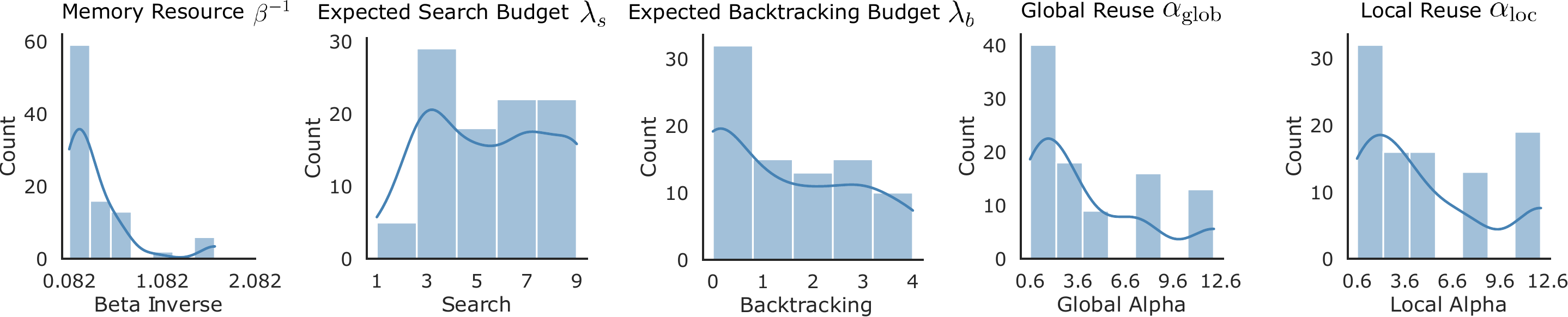}
    \caption{\textbf{Empirical distribution of fitted HAG parameters.} 
    Histograms show the distribution of participant-level maximum-likelihood estimates for the five HAG parameters. Curves indicate kernel density estimates to visualize central tendency and dispersion. The distributions reveal substantial between-participant variability.
    }
    \label{fig:si:fitted-param-distribution}
\end{figure}